%% file: main.tex
\documentclass[authoryear]{elsarticle}

\makeatletter
\nopreprintlinetrue
\makeatother

\usepackage{bbm}
\usepackage[table]{xcolor}
\usepackage{graphicx}
\usepackage{soul, color}
\usepackage{rotating}
\usepackage{multirow}
\usepackage{booktabs}
\usepackage{amsfonts}
\usepackage{amsmath}
\usepackage{amssymb}
\usepackage{url}
\usepackage{colortbl}
\usepackage{lipsum}
\usepackage{xspace}

\usepackage{pifont}

\newcommand{\cmark}{\ding{51}} 
\newcommand{\xmark}{\ding{55}} 

\usepackage{xcolor}
\definecolor{quartobg}{HTML}{F7F7F7}
\usepackage{minted}
\usepackage{caption}
\definecolor{codeorange}{rgb}{0.9,0.3,0.}
\definecolor{codeblue}{rgb}{0.,0.5,0.99}
\definecolor{codepurple}{rgb}{0.58,0,0.82}
\definecolor{codegreen}{rgb}{0.,0.5,0.}
\definecolor{backcolour}{rgb}{0.95,0.95,0.92}
\definecolor{backcolour2}{rgb}{0.96,0.96,0.96}
\definecolor{codeblack}{rgb}{0.,0.,0.}

\newcommand{\Prob}{\mathbb{P}}
\newcommand{\rmNB}{\mathrm{NB}}
\newcommand{\rmPois}{\mathrm{Pois}}
\newcommand{\rmBeta}{\mathrm{Beta}}
\newcommand{\rmGamma}{\mathrm{Gamma}}
\newcommand{\rmBer}{\mathrm{Ber}}

\newcommand{\pkg}[1]{{\normalfont\fontseries{b}\selectfont #1}} 
\newcommand{\code}[1]{\texttt{#1}} 
\newcommand{\proglang}[1]{\textsf{#1}} 

\newif\ifblind
\blindfalse

\newcommand{\fableintermittentname}{%
  \ifblind fable.BLINDED\else fable.intermittent\fi
}

\newcommand{\fableintermittent}{\pkg{\fableintermittentname}\xspace}
\newcommand{\tweediedistr}{%
  \ifblind
    \pkg{tweedieBLINDED}%
  \else
   \pkg{tweedieDistr}%
  \fi
  \xspace
}

\begin{document}

\begin{frontmatter}
\title{\fableintermittent: benchmarking probabilistic forecasting methods for intermittent time series}

\ifblind
\else

\author[1]{Stefano Damato\corref{cor1}}
\ead{stefano.damato@supsi.ch}

\author[1]{Lorenzo Zambon}
\ead{lorenzo.zambon@supsi.ch}

\author[1]{Giorgio Corani}
\ead{giorgio.corani@supsi.ch}

\author[1]{Dario Azzimonti}
\ead{dario.azzimonti@supsi.ch}

\cortext[cor1]{Corresponding author}

\affiliation[1]{organization={SUPSI, Istituto Dalle Molle di Studi sull'Intelligenza Artificiale (IDSIA)},
city={Lugano},
country={Switzerland}}
\fi

\begin{abstract}
Intermittent time series are common in spare-parts demand and retail sales.
Since the cost of forecast errors is typically asymmetric, decisions such as inventory control require the full predictive distribution rather than a point forecast.
Many probabilistic forecasting methods have been proposed;
their implementations, however, are scattered across different software frameworks, 
making it difficult to compare them systematically.
We introduce \fableintermittent, an \proglang{R} package that implements several probabilistic forecasting methods for intermittent series within the \pkg{fable} framework. 
The package allows several models to be fitted and evaluated on a collection of time series through a single, simple forecasting pipeline.
We also introduce TWEES, a new exponential smoothing model with a Tweedie predictive distribution.
Fitting TWEES requires repeated evaluation of the computationally demanding Tweedie density. We also release the \proglang{R} package \tweediedistr, whose implementation of the Tweedie distribution  is substantially faster than the existing one while preserving the same numerical accuracy.
We evaluate the methods implemented in \fableintermittent on four datasets, also released in the package.
\end{abstract}

\end{frontmatter}

\newpage

\section{Introduction}

Intermittent time series are non-negative valued time series characterized by positive values interspersed with zeros \citep{Boylan_Syntetos_2021}.
They commonly arise, for instance, in supply-chain applications \citep{Syntetos_Babai_Boylan_Kolassa_Nikolopoulos_2016} and retail \citep{Fildes_Kolassa_Ma_2022}. 
Most methods for intermittent time series \citep{Croston_1972, Syntetos_Boylan_2005, Teunter_Syntetos_Babai_2011}
provide point forecasts, which are insufficient to support decision making. The costs of forecasting errors are usually asymmetric, so that the optimal forecast is a quantile rather than the mean of the predictive distribution \citep{Kolassa_2016};
inventory control, for instance, requires the upper tail of the predictive distribution of
the demand \citep{Prak_Teunter_2019}.

Probabilistic models for intermittent time series span several methodologies, from statistical to machine learning approaches \citep{Lang_Mao_He_Deng_Liu_Zuo_2026}: static empirical and parametric distributions \citep{Boylan_Babai_2022,Kolassa_2016}, bootstrap methods \citep{Willemain_Smart_Schwarz_2004,Zhou_Viswanathan_2011}, Bayesian dynamic models \citep{Harvey_Fernandes_1989, Babai_Chen_Syntetos_Lengu_2021}, exponential smoothing models for count data \citep{Snyder_Ord_Beaumont_2012, Svetunkov_Boylan_2023}, ARMA-based models \citep{Sbrana_2025,Sbrana_Babai_2026}, Gaussian processes for intermittent data \citep{Damato_Azzimonti_Corani_2025}. 
Probabilistic forecasts can also be produced by a global deep-learning model such as
DeepAR \citep{Salinas_Flunkert_Gasthaus_Januschowski_2020},
trained with a distribution head suitable for intermittent data. 

Only a few methods, however, have publicly available implementations. 
In \proglang{R} \citep{R_language}, the \pkg{smooth} package \citep{smoothpkg} implements the intermittent exponential smoothing (iETS) model \citep{Svetunkov_Boylan_2023};
while \pkg{fableCount} \citep{fable_count} makes available the models GLARMA and INGARCH for time series of counts. 
In \proglang{Python}, \pkg{GluonTS} \citep{gluonts_jmlr}  provides global deep-learning models. 
Other models, e.g. \citet{Damato_Azzimonti_Corani_2025} and \citet{Sbrana_Babai_2026},  are only released in project repositories. 
Available implementations, moreover, are spread across different frameworks, making it laborious to compare several probabilistic methods for intermittent time series on the same data.

We fill this gap with \fableintermittent, an \proglang{R} package which extends the \pkg{fable} framework \citep{fable_ohara} by implementing several probabilistic forecasting methods from the literature. The package allows these methods to be fitted and evaluated on collections of time series with a straightforward pipeline.

\fableintermittent also provides TWEES, a novel probabilistic model for intermittent time series. 
It is an exponential smoothing model with a Tweedie predictive distribution.
The Tweedie distribution is suitable for intermittent data as it combines a point mass at zero with a continuous mixture of Gamma distributions over the positive real line. Recent studies \citep{Damato_Azzimonti_Corani_2025,damato2026_local_global} show that it improves the estimation of the highest quantiles compared with distributions traditionally used for forecasting of intermittent time series, such as the negative binomial and hurdle-shifted negative binomial.
Following the design of \citet{Snyder_Ord_Beaumont_2012}, we use two different exponential smoothing recursions to model the mean of the predictive distribution and the probability of demand occurrence. In our experiments, TWEES is among the most accurate models.

The evaluation of the Tweedie density, however, is computationally demanding \citep{Dunn_Smyth_2005}. This is problematic when the Tweedie distribution is the predictive distribution of a time series model, as the density must be evaluated repeatedly during estimation, especially when the model is fitted on a large collection of time series. We address this limitation with \tweediedistr, an \proglang{R} package providing a novel implementation of the Tweedie distribution. For density evaluation, our implementation is up to 20 times faster than that of the \pkg{tweedie} package \citep{tweedie_pkg}, while retaining the same numerical accuracy.
Both \fableintermittent and \tweediedistr are released under the LGPL-3.0 License\footnote{\url{https://gnu.org}} and available on CRAN\footnote{
\ifblind
\url{https://cran.r-project.org}. The packages links will be provided after acceptance. 
\else
\url{https://cran.r-project.org/web/packages/fable.intermittent/index.html}, \\ \url{https://cran.r-project.org/web/packages/tweedieDistr/index.html}.
\fi}, the official archive of \proglang{R}
packages.

The paper is organized as follows. Sec.~\ref{sec:fableintermittent} presents the models 
implemented in \fableintermittent, the datasets available within the package, and an example of the forecasting pipeline.
Sec.~\ref{sec:tweedie} describes our implementation of the Tweedie distribution, available in \tweediedistr, and its speedup over the implementation provided by \pkg{tweedie}.
Sec.~\ref{sec:twees} introduces the TWEES model.
Sec.~\ref{sec:experiments} evaluates the models of \fableintermittent on several datasets.
Sec.~\ref{sec:integration} shows how the resulting forecasts can be probabilistically reconciled using \pkg{fable.bayesRecon} \citep{fable_bayesRecon}.
Finally, Sec.~\ref{sec:conclusions} presents our conclusions.

\section{Literature models and datasets provided by \fableintermittent}
\label{sec:fableintermittent}
\input{tables/methods_table}

\subsection{Models}\label{sec:models}

The package implements the probabilistic models for intermittent time series listed in Tab.~\ref{tab:methods}.
The last column indicates which models rely on \emph{Croston's decomposition} \citep{Croston_1972}, which decomposes the series into an occurrence process, indicating whether demand is positive, and a demand size process, containing the positive demand values.
According to the model classification by \citet{Januschowski_Gasthaus_Wang_Salinas_Flunkert_Bohlke-Schneider_Callot_2020}, all the implemented models are local.

Bootstrapping methods for intermittent demand, reviewed by \citet{Hasni_Babai_Aguir_Jemai_2019}, resample past demand values to estimate the demand size process. In particular, we implement
WSS \citep{Willemain_Smart_Schwarz_2004}, which models the occurrence process  as a two-state Markov Chain, and VZ \citep{Zhou_Viswanathan_2011}, which bootstraps demand intervals. The Bayesian methods of \citet{Harvey_Fernandes_1989} use a prior for the parameters of the predictive distribution, and update it as the time series evolves; GAMPOISB uses a Gamma prior for the Poisson parameter, while BETANBB uses a Beta prior for the probability parameter of a negative binomial model, which count parameter is learned via maximum likelihood. \citet{Snyder_Ord_Beaumont_2012} couple exponential smoothing with predictive distributions suitable for count data: 
in HSPES, two processes smooth the rate of a Poisson on the demand size and a probability of occurrence to get a hurdle-shifted Poisson forecast distribution, while in NEGBINES a single process determines the mean of a negative binomial distribution.

ARMA-based models use a Gaussian likelihood, but the mean is tailored to properties of intermittent demand: \citet{Sbrana_2025} additionally models the occurrence with a Markov chain, while \citet{Sbrana_Babai_2026} constrain the mean of the forecast distribution to be positive.
Finally, static distributions 
treat time series data as i.i.d. observations; they are considered a solid benchmark for intermittent time series \citep{Spiliotis_Makridakis_Kaltsounis_Assimakopoulos_2021}. We implement EMPSD, the simple empirical distribution of the data \citep{Boylan_Babai_2022}, and PARAMSD that fits a parametric distribution \citep{Kolassa_2016}. Our implementation fits five candidate parametric distributions (Poisson, negative binomial, their hurdle-shifted versions and a discretized Tweedie distribution) and selects among them by minimizing the Bayes Information Criterion \citep[BIC,][]{schwarz1978estimating}. See~\ref{app:methods} for more detailed descriptions.

Some of the methods above share the same routines, which are run several times for each time series fit. Our package improves the computational efficiency of these methods by implementing the repeated routines, such as exponential smoothing and ARMA recursions and Bayesian updates, in \proglang{C++} via the package \pkg{Rcpp} \citep{rcpp_2011}.



\subsection{Datasets}\label{sec:datasets}

\input{tables/datasets}

The package provides four datasets of intermittent time series (Tab.~\ref{tab:datasets}) in a format ready to be ingested by the \pkg{fable} pipeline. 
\emph{Auto} \citep{turkmen2021forecasting} contains monthly demand for automotive spare parts; its series are short and only mildly intermittent. \emph{Pasta} \citep{mancuso2021machine} contains daily sales of 118 pasta products, including promotion covariates. \emph{RAF} \citep{Syntetos_Boylan_2005} contains monthly demand for spare parts of the Royal Air Force.
\textit{TinyM5} is a subset from the data set of the M5 competition \citep{Makridakis_Spiliotis_Assimakopoulos_2022} and contains daily sales of different products at Walmart; we use the selection of \citet{m5package}. Both \emph{Pasta} and \emph{RAF} are characterized by demand spikes (high $\mathrm{CV}^2$), which challenge the tails of the predictive distributions. \emph{RAF} has also the most intermittent demand. 

\begin{figure}[h!]
\centering
\includegraphics[width=\textwidth]{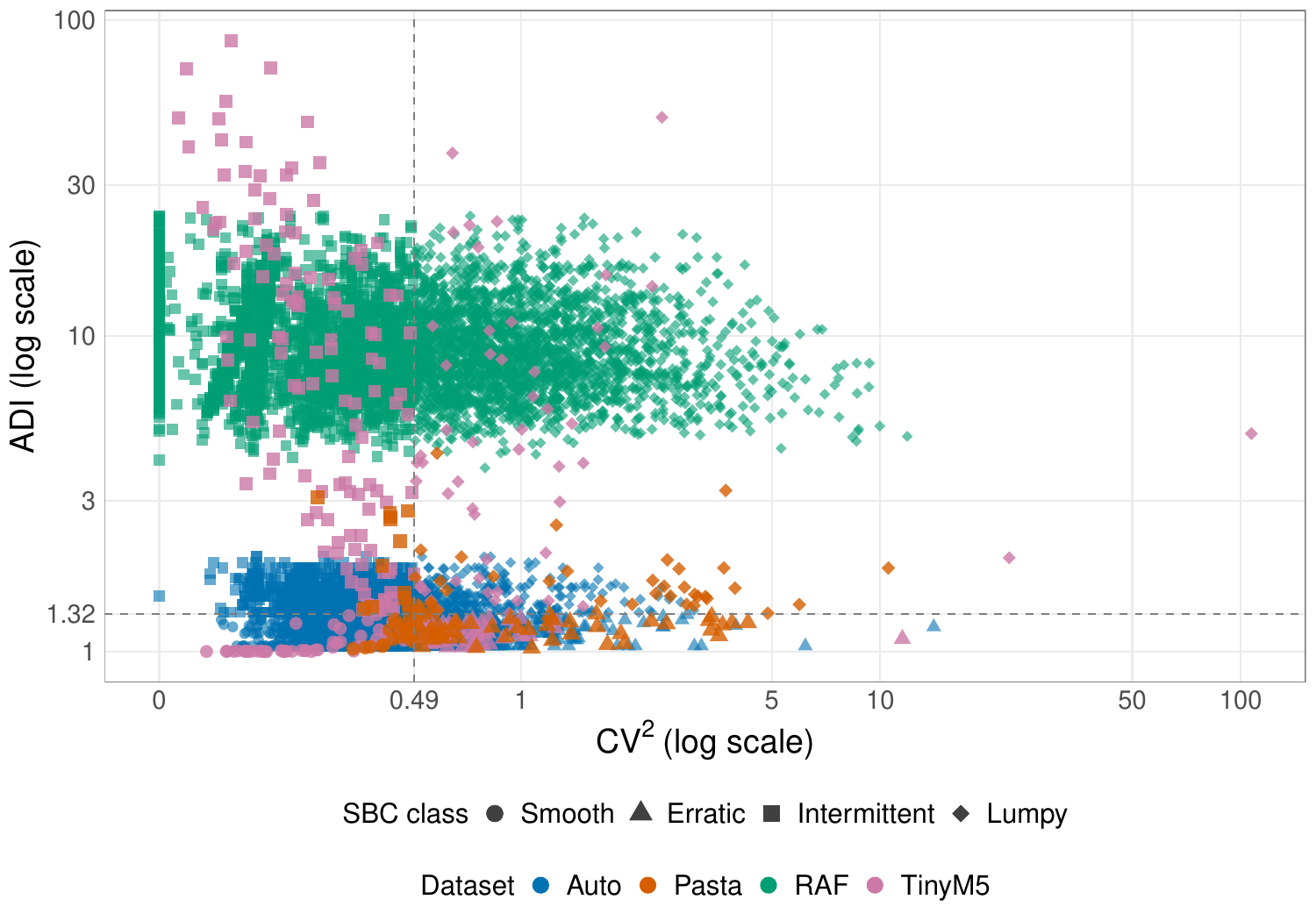}
\caption{Classification of the series of the four data sets according to average inter-demand interval (ADI, log scale) and the
squared coefficient of variation of the non-zero demand sizes (CV$^2$, linearly-adjusted log
scale). Dashed lines mark the cut-offs of
\citet{syntetos2005categorization} ($\mathrm{ADI}=1.32$,
$\mathrm{CV}^2=0.49$). A small number of series with $\mathrm{CV}^2 = 0$
(constant non-zero demand size) appear in the plot exactly on zero because of the linearly adjusted log scale.}
\label{fig:sbc}
\end{figure}

Fig.~\ref{fig:sbc} shows the distribution of the average inter-demand interval (ADI) and the squared coefficient of variation of non-zero demand sizes (CV$^2$) across the time series of the different datasets, following the classification of \citet{syntetos2005categorization}.

The data sets are released in the \pkg{tsibble} \citep{tsibble_wang} format, a tidy data structure to store time series. In each data set, the column \code{value} stores the actual observations, and the column \code{index} saves their timestamps. One or more additional columns serve as time series identifiers; others may provide exogenous covariates. 

\subsection{Easy benchmarking with \fableintermittent}

The package \fableintermittent extends \pkg{fable}
\citep{fable_ohara}, the tidy time series modelling framework
of the \textit{tidyverts}\footnote{\url{https://tsibble.tidyverts.org/}} ecosystem.
All models share a common syntax and can be straightforwardly fitted to collections of time series stored as \pkg{tsibble} objects. Their probabilistic forecasts are represented as \pkg{distributional} objects \citep{distributional_ohara}. This common representation allows the models to be evaluated through the same pipeline, based on \pkg{fabletools} \citep{fabletools} functions, and compared with other models available in \pkg{fable} or its extensions.
Listings~\ref{lst:benchmark1} and~\ref{lst:benchmark}  show the basic pipeline to obtain a tsibble, \code{res}, with the results of Sec.~\ref{subsec:results}. 
In a few lines, we train and evaluate 11 models, the ten models of Tab.~\ref{tab:methods} and the new model TWEES introduced in Sec.~\ref{sec:twees}. 

\begin{listing}[h!]
\begin{minted}[
    linenos,
    numbersep=5pt,
    style=tango,
    bgcolor=quartobg,
    fontsize=\small,
    breaklines,
    frame=none,
    xleftmargin=1.5em,  % keeps numbers from crowding the background box edge
    escapeinside=@@
]{r}
library(fable)
library(@\fableintermittentname@)
library(dplyr)

# forecast 28 steps ahead
h <- 28

# Create train/test split of pasta data set
train <- pasta |> filter(index <= max(index) - h)
test  <- pasta |> filter(index >  max(index) - h)
head(train,1)

# A tsibble: 1 x 5 [1D]
# Key:       brand, product [1]
  index      brand product value promotion
  <date>     <chr> <chr>   <dbl>     <int>
1 2014-01-02 B1    1           7         0
\end{minted}
\caption{Load the \code{pasta} data set and create a train/test split. Lines 12-16 show the shape of the tsibble \code{train} where the column \code{value} contains the values of the time series.}
\label{lst:benchmark1}
\end{listing}

\begin{listing}[h!]
\begin{minted}[
    linenos,
    numbersep=5pt,
    style=tango,
    bgcolor=quartobg,
    fontsize=\small,
    breaklines,
    frame=none,
    xleftmargin=1.5em  % keeps numbers from crowding the background box edge
]{r}
# Fit models on train
fit <- train |>
    model(
      wss       = WSS(value),
      vz        = VZ(value),
      betanbb   = BETANBB(value),
      gampoisb  = GAMPOISB(value),
      hspes     = HSPES(value),
      negbines  = NEGBINES(value),
      twees     = TWEES(value),
      empsd     = EMPSD(value),
      paramsd   = PARAMSD(value),
      marwal    = MARWAL(value),
      nnarma    = NNARMA(value)
    )
    
# Forecast 28 steps ahead
fc <- fit |>
    forecast(h = h)

# compute the accuracy, group by model and average over the dataset.
res <- fc |>
    accuracy(data, measures = experiment_measures) |>
    group_by(.model) |>
    summarise(across(where(is.numeric), \(x) mean(x, na.rm = TRUE)))
\end{minted}
\caption{The pipeline for fitting, forecasting and evaluation on the Pasta dataset. Here, \code{experiment\_measures} is a list of \pkg{fabletools} forecast evaluation measures; \code{.model} is the column where model names are stored, over which we group the results and compute the mean.}
\label{lst:benchmark}
\end{listing}

\section{\tweediedistr: a fast implementation of the Tweedie distribution}
\label{sec:tweedie}

\subsection{The Tweedie distribution}
\label{sec:tweediedistr}

\begin{figure}[h!]
\centering
\includegraphics[width=\textwidth]{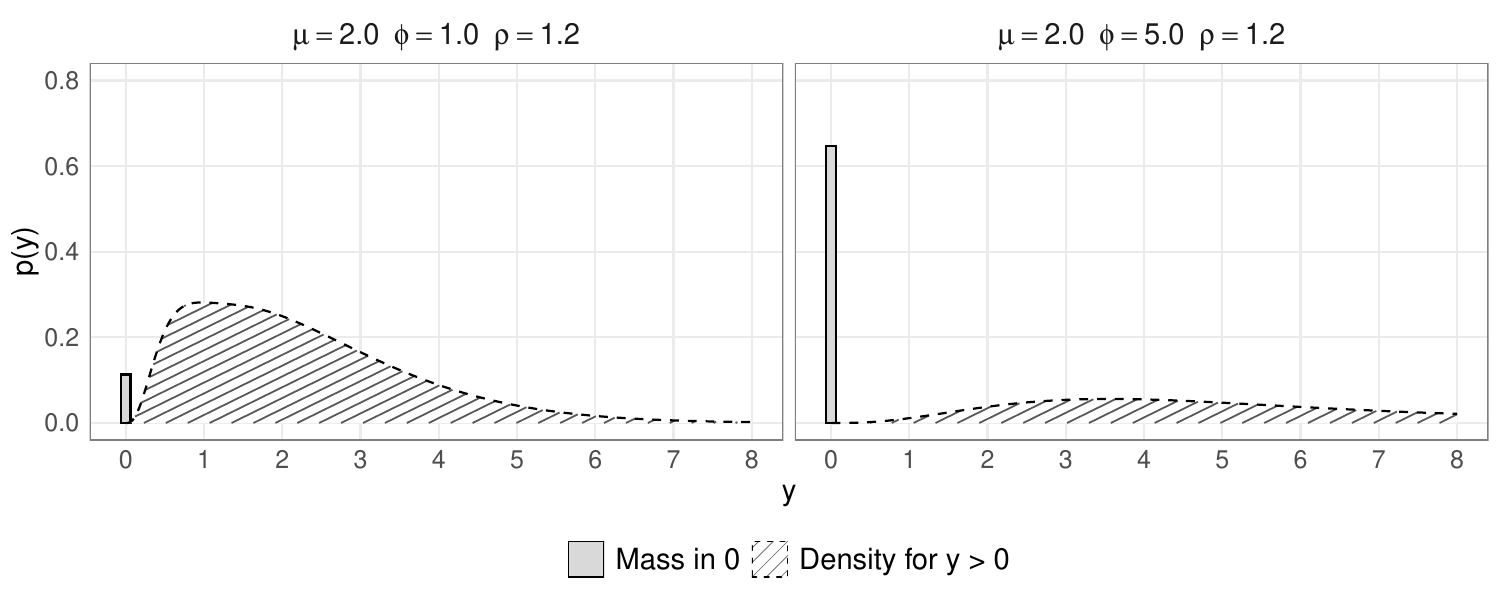}
\caption{Two Tweedie distributions with the same mean and power parameter, but different dispersion.}
\label{fig:tweedie}
\end{figure}

The Tweedie is a family of exponential dispersion distributions \citep{Jrgensen1987ExponentialDM} characterized by a power mean-variance relationship. If $Y \sim \mathrm{Tw}(\mu, \phi, \rho)$, then
\begin{equation}\label{eq:tweedie_var}
 \mathrm{Var}(Y) = \phi \mu^\rho,
\end{equation}
where $\mu>0$ is the mean, $\phi>0$ is the dispersion parameter, and $\rho>0$ is the power parameter.
As in \citet{Damato_Azzimonti_Corani_2025}, we limit the power to $\rho \in (1,2)$: for this choice of the power parameter, the Tweedie distribution can be parameterized as a Compound Poisson Gamma model:
\begin{align}
\label{eq:GammaPois}
Y = \sum_{i=1}^N X_i, \quad \text{with} \\ \nonumber
X_i \overset{i.i.d.}{\sim} \mathrm{Gamma}(\alpha, \beta), \quad
N \sim \mathrm{Pois}(\lambda),
\end{align}
where $\lambda$, $\alpha$, and $\beta$ are functions of the original  parameters \citep{Dunn_Smyth_2005}. 
From this parametrisation it can be seen why the Tweedie is suitable for intermittent demand: the sum of Gamma random variables models the positive mass and allows for a flexible right tail behaviour, while the Poisson distribution provides an atomic mass in zero computed from Eq.~\eqref{eq:GammaPois} as 
\begin{equation}\label{eq:lambda}
  P(Y=0) = P(N=0) = e^{-\lambda} = e^{-\frac{\mu^{2 - \rho}}{\phi ( 2 - \rho)}}
\end{equation}
Fig.~\ref{fig:tweedie} shows two Tweedie distributions with the same mean ($\mu=2$) and power ($\rho=1.2$), but different dispersion $\phi$;
this results in 
different probabilities of zero and  different distributions of positive values.

\subsection{A novel, fast implementation in \tweediedistr} 

For $y > 0$, the evaluation of the Tweedie density is computationally cumbersome because it requires computing infinite sums. 
The \proglang{R} package \pkg{tweedie} \citep{tweedie_pkg} uses
evaluation strategies based on series expansion \citep{Dunn_Smyth_2005} and Fourier inversion of the characteristic function \citep{Dunn2008}.
The running times of these implementations, however, can become a bottleneck when fitting a Tweedie distribution to each time series of a large collection, or when the Tweedie distribution is used as the likelihood of a time series model and must therefore be evaluated repeatedly.

Our package \tweediedistr provides a substantially faster implementation of the Tweedie distribution. Similarly to the package of \citet{tweedie_pkg}, \tweediedistr follows the \proglang{R} statistical convention implementing functions for the evaluation of the density function (\code{dtweedie()}), the  cumulative distribution function (\code{ptweedie()}) and its inverse (\code{qtweedie()}), and for random sample generation (\code{rtweedie()}). 

\begin{figure}[!h]
  \centering
  \ifblind
  \includegraphics[width=1\linewidth]{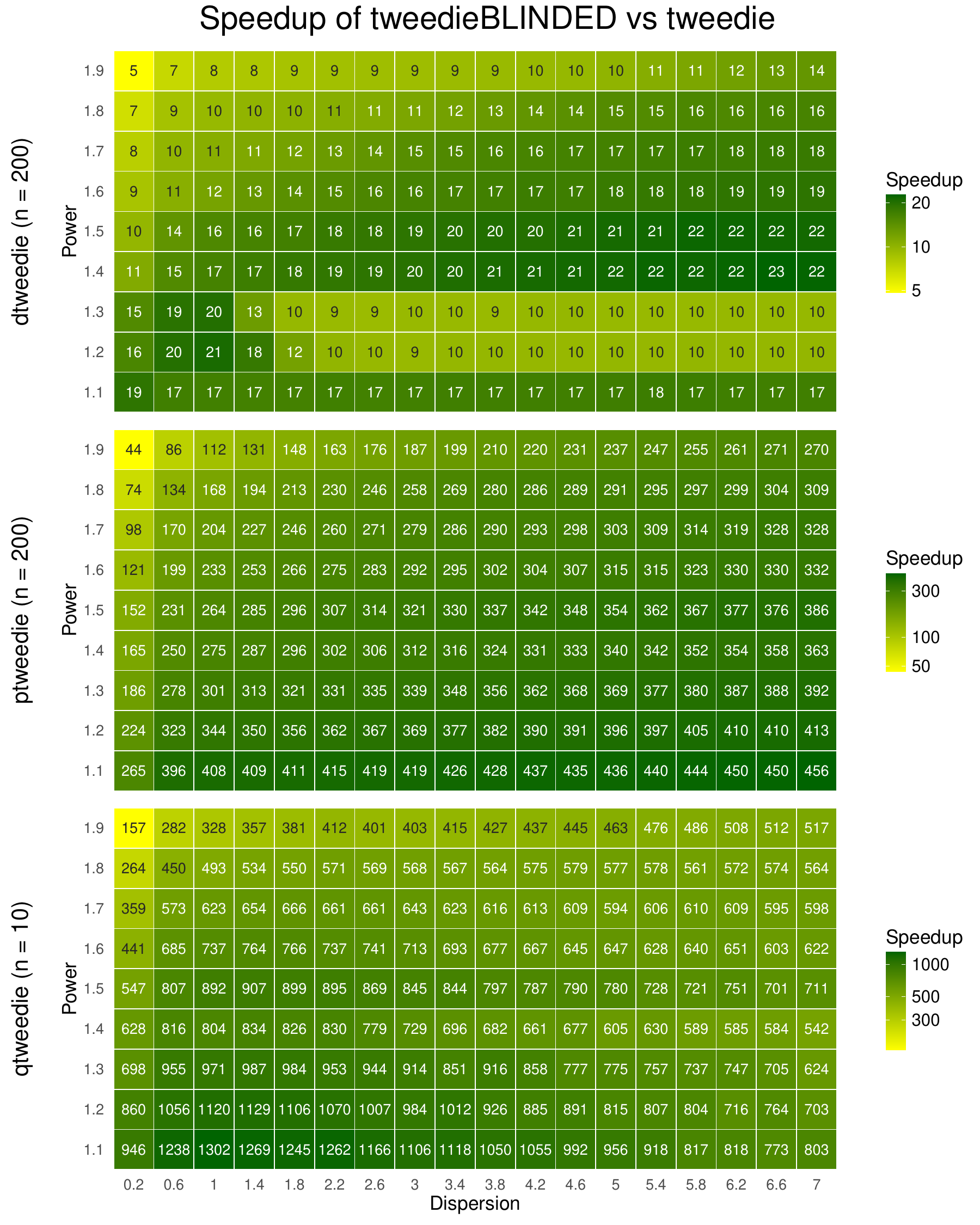}
  \else \includegraphics[width=1\linewidth]{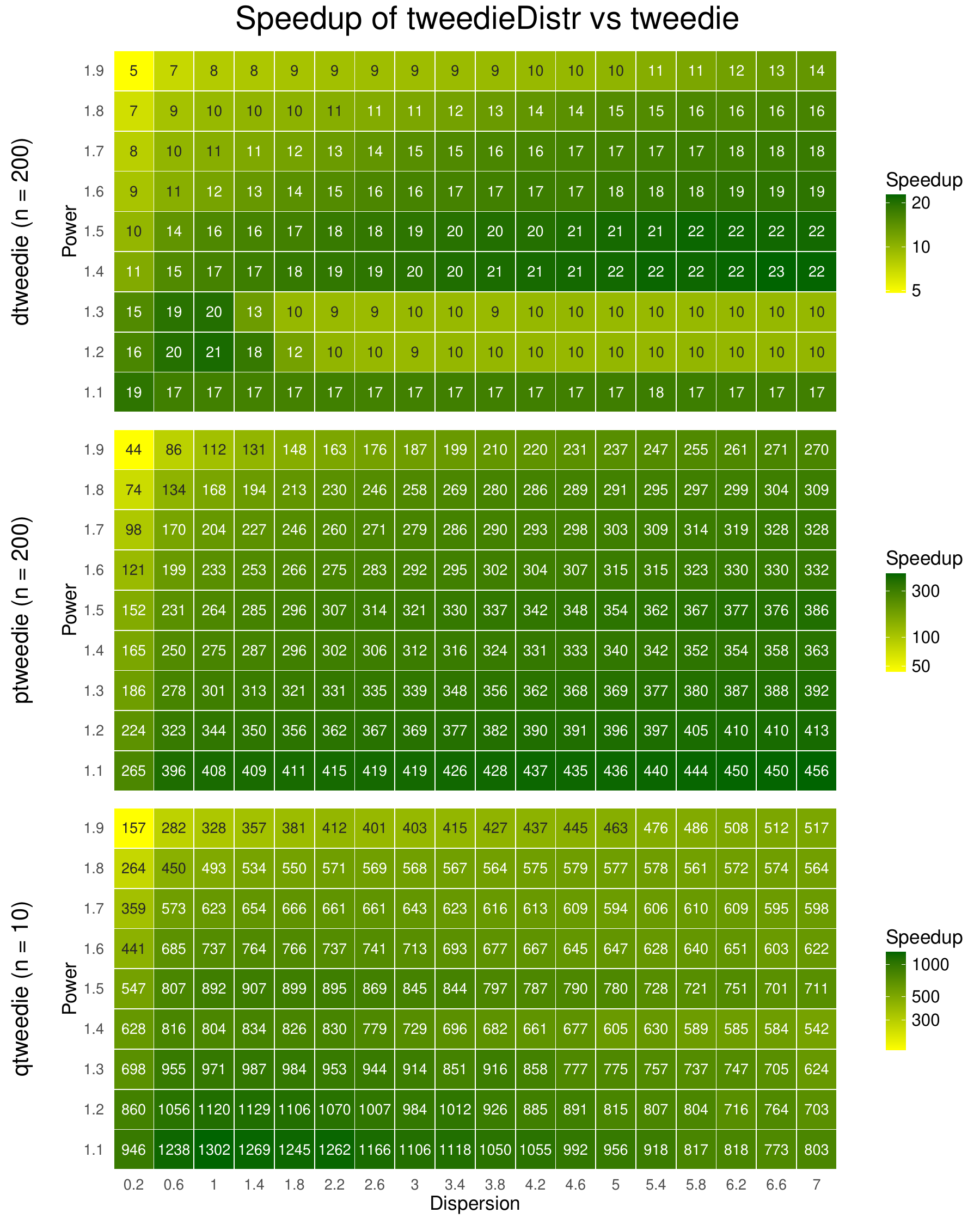}
  \fi
  \caption{Improvement of the running times of the core density (\code{dtweedie()}), cumulative (\code{ptweedie()}) and quantile (\code{qtweedie()}) functions for different choices of $\phi$ and $\rho$. For additional details on the setup of this evaluation, see~\ref{appendix:tweedie_exp}.}
  \label{fig:tweedie_speed}
\end{figure}

In Fig.~\ref{fig:tweedie_speed} we show the speedup on the three core functions \code{dtweedie()}, \code{ptweedie()}, and \code{qtweedie()} provided by \tweediedistr over \pkg{tweedie} 
for different values of power, dispersion, and sample size. See \ref{appendix:tweedie_exp} for more details about our experimental setup. 

The implementations of \tweediedistr are on average 15, 300, and 600 times faster than their counterpart in \pkg{tweedie}. 
The speedup can be attributed, in part, to the use of \proglang{C++} via \pkg{Rcpp}/\pkg{RcppArmadillo} \citep{rcpp_2011} in our implementations. Moreover, a substantial advantage is provided by different choices in terms of algorithms, detailed in \ref{appendix:tweedie_algorithms}. 
Finally, \tweediedistr vectorises over mean, dispersion
and power, whereas \pkg{tweedie} requires an \proglang{R}-level loop for the power parameter, as it only treats it as a scalar value. 

The function \code{dtweedie()} from \tweediedistr achieves the same accuracy as the one in \pkg{tweedie}. Instead, the cumulative function evaluation (\code{ptweedie()}) is 2\% more accurate in \tweediedistr than in \pkg{tweedie} on average, when compared to Monte Carlo evaluations. Moreover, the \pkg{tweedie} implementation of the inverse  cumulative distribution (\code{qtweedie()}) sometimes returns an error when $\rho$ approaches 2. We observed such failures in the upper row of the third panel in Fig.~\ref{fig:tweedie_speed}; they are more frequent in cells closer to the top left corner, that is, when $\phi$ grows larger.  However, the scope of the \pkg{tweedie} package is wider, as it implements the functions above for any $\rho > 0$, while \tweediedistr currently handles only $\rho \in (1, 2)$.

\section{The TWEES model}\label{sec:twees}
TWEES is an exponential smoothing model with a Tweedie predictive distribution.
A smoothed level controls the mean of a Tweedie likelihood, 
following a simple exponential smoothing recursion as in \citet{Snyder_Ord_Beaumont_2012}. Thus, denoting the data of time series as $y_1, \dots, y_T$, TWEES models $Y_i \sim \mathrm{Tw}(\mu_i, \phi_i, \rho)$, whit the recursion
\[
\mu_i = \alpha_\mu y_{i-1} + \theta_\mu \bar{\mu} + (1 - \alpha_\mu - \theta_\mu) \mu_{i-1}
\]
where $\alpha_\mu$ and $\theta_\mu$ are learnable smoothing parameters, and $\bar{\mu}$ is the average of the training data.
This model is also similar to the TweedieGP model of \citet{Damato_Azzimonti_Corani_2025},
where the mean of the Tweedie distribution is modeled with a Gaussian process.
Recall that the variance of $Y_i$ is tied to its mean
through Eq.~\eqref{eq:tweedie_var}; here, however, the dispersion $\phi_i$ is not kept fixed, but derived from a separate occurrence process, described next.

To let occurrence and demand size evolve independently, we run a second damped exponential smoothing on the occurrence indicator $o_t = \mathbbm{1}_{y_t > 0}$, as in the
hurdle-shifted Poisson model of \citet{Snyder_Ord_Beaumont_2012}:
\[
  \pi_i = \alpha_\pi o_{i-1} + \theta_\pi \bar{\pi} + (1 - \alpha_\pi - \theta_\pi) \pi_{i-1},
\]
where $\alpha_\pi$ and $\theta_\pi$ are smoothing parameters too, and $\bar{\pi} $ is the average of the occurrence process. Then, we constrain $\phi_i$ so that the mass at zero of the Tweedie matches the occurrence forecast
$\pi_i$: since the zero-mass of a Tweedie variable is $e^{-\lambda_i}$ with $\lambda $ as in Eq.~\eqref{eq:lambda}, setting
$e^{-\lambda_i} = 1 - \pi_i$ and solving for $\phi_i$ gives
\[
\phi_i = \frac{\mu_i^{2-\rho}}{(2-\rho)\left(-\log(1-\pi_i)\right)}.
\]

The parameters $(\rho, \mu_0, \alpha_\mu, \theta_\mu, \pi_0, \alpha_\pi, \theta_\pi)$ are
estimated jointly by maximising the Tweedie log-likelihood implied by the recursions
above, subject to $\theta_\eta \ge 0, \alpha_\eta \ge 0$, and the usual
smoothing constraints $\alpha_\eta + \theta_\eta < 1$ for $\eta \in \{\mu, \pi\}$.

One-step-ahead forecasts are obtained directly from the fitted model, given the observed values; longer-horizon forecasts are generated by simulating both exponential smoothing processes forward in time.

As in \citet{Damato_Azzimonti_Corani_2025}, the series is rescaled by its median positive value before fitting; this operation is allowed as the distribution is absolutely continuous for $y > 0$.
In TWEES we restrict the power parameter to $\rho \in (1.2, 1.8)$: 
the upper bound limits the number of terms required to evaluate the density, while the lower bound ensures numerical stability. 

\section{Experiments}
\label{sec:experiments}

\subsection{Setup}

We evaluate the models introduced in Sec.~\ref{sec:models} and~\ref{sec:twees} on the datasets introduced in Sec.~\ref{sec:datasets}. We use \fableintermittent~(v. 0.3.0) and \tweediedistr~(v. 0.2.0); our code is available in this anonymous GitHub repository\footnote{\url{https://anonymous.4open.science/r/benchmarking_intermittent-DD78}; note that in this repository, the name of our packages is no longer anonymous.}. All models are run with default options; for those requiring sample paths, $10^5$ are drawn.


For each dataset, we perform forecast evaluation using an expanding window approach where we use the first $T$ observations as a training set and the following $h$ observations as a holdout test set. In particular, we use two windows setting $T = L - j h$ for $j = 1, 2$. The value of
$h$ is set to the dataset's native forecast
horizon: $h=6$ for Auto, $h=12$ for RAF, and $h=28$ for Pasta and TinyM5.

Writing the training set by $y_1,  \allowbreak \dots, \allowbreak y_T$ and the holdout test set by $y_{T+1}, \dots, \allowbreak y_{T+h}$, we use the Root Mean Squared Scaled Error \citep{hyndman2006accuracy} to evaluate point forecasts:

$$
\mathrm{RMSSE} = \sqrt{\dfrac{\dfrac{1}{h}\sum_{t=T+1}^{T+h}\left(y_t-\hat y_t\right)^2}
{\dfrac{1}{T-1}\sum_{t=2}^{T}\left(y_t-y_{t-1}\right)^2}}.
$$
where $\hat{y}_t$ denotes the mean of the forecast distribution at time $t$.
We assess the predictive distribution using the quantile score \citep{Gneiting_Raftery_2007}
$$
\mathrm{QS}_\tau(y, \hat{y}_\tau) =
\begin{cases}
\tau\,(y - \hat{y}_\tau) & \text{if } y \ge \hat{y}_\tau, \\
(1-\tau)\,(\hat{y}_\tau - y) & \text{if } y < \hat{y}_\tau.
\end{cases}
$$
Following \citet{Spiliotis_Makridakis_Kaltsounis_Assimakopoulos_2021}, we evaluate the quantile levels $\tau = 0.5, 0.75, \allowbreak 0.835, \allowbreak 0.975, 0.995$, and we scale the quantile scores by the Mean Absolute Error (MAE) of the naive forecast. We omit lower-tail levels ($\tau=0.005,0.025, \allowbreak 0.165, \allowbreak 0.25$), as in the data sets considered their quantile forecast is zero for most time series and models. The resulting scaled quantile score $\mathrm{sQS}_\tau$ is defined as follows:
$$
\mathrm{sQS}_\tau = \frac{\sum_{t = T+1}^{T+h}\mathrm{QS}_\tau(y_t, \hat{y}_{t,\tau})}{
\sum_{t=2}^T | y_t - y_{t-1} |
},
$$
 where $\hat{y}_{t,\tau}$ is the quantile forecast of level $\tau$ at timestamp $t$.
We note that $\mathrm{sQS}_{0.5}$ corresponds to the Mean Absolute Scaled Error \citep{hyndman2006accuracy}.

The scoring rules we introduced are evaluated over the test set of each window: subsequently, we compute a score for each time series by taking the mean over the two windows. 
Since the scoring rules are scale-independent, we evaluate each method on each dataset by taking the mean across different time series; this practice has been shown to be more robust than the use of ranks with MCB test \citep{KONING2005397}, especially when dealing with high quantile levels \citep{corani2026modelselectionproperscoring}. To test the significance of our findings, we use, for each scoring rule, the Model Confidence Set
(MCS, \citealp{MCS_hansen_2011}; confidence $\alpha = 0.05$), a procedure that identifies a set of statistically equivalent best methods by iteratively eliminating the worst one. 

\subsection{Results}
\label{subsec:results}

\input{tables/auto_spl_MAE}
\input{tables/pasta_spl_MAE}
\input{tables/raf_spl_MAE}
\input{tables/tinym5_spl_MAE}

We report the average scores of each method in Tab.~\ref{tab:auto_spl_MAE} (Auto), Tab.~\ref{tab:pasta_spl_MAE} (Pasta), Tab.~\ref{tab:raf_spl_MAE} (RAF), and Tab.~\ref{tab:tinym5_spl_MAE} (TinyM5).
For each metric, we report in bold the methods with the lowest score (the best), we underline the scores of those methods falling into the MCS, and we report the rank of each method in grey.

Across all four datasets, no single method dominates uniformly, but two patterns can be observed. First: static, simple benchmarks like EMPSD and PARAMSD are hard to beat on RMSSE and on central quantiles.
As already noted by \citet{Boylan_Babai_2022}, static methods are particularly effective when the historical data is limited (Auto, Tab.~\ref{tab:auto_spl_MAE}) or very sparse (RAF, Tab.~\ref{tab:raf_spl_MAE}). The difference between the two methods is often minimal, except on the Auto dataset, where the parametric model is better than the empirical distribution on all metrics.

Second, we observe that two methods, TWEES and NEGBINES, have better scores and almost never fall outside the top five. For instance, on the TinyM5 dataset, Tab.~\ref{tab:tinym5_spl_MAE}, the two methods are always part of the MCS. TWEES, in particular, ranks in the top three on RMSSE and $\mathrm{sQS}_{0.835}$ across all datasets. 

Our results show that exponential smoothing models generally perform better. For TWEES, in particular, the two-process approach for occurrence and demand size is key: using a single process that only controls the mean, as in NEGBINES, led to poorer results in preliminary experiments. The negative binomial method, however, remains very competitive, and almost always ranks better than TWEES on Pasta, Tab.~\ref{tab:pasta_spl_MAE}. HSPES is the worst of the three ES methods: indeed, its modeling of the occurrence is effective, but the demand process is fit on comparatively few non-zero observations and the light right tail of its Poisson likelihood struggles to cover the highest quantiles.

A related problem affects the GAMPOISB model: the Gamma distribution driving the mean of the Poisson often has low variance, constraining the model to behave similarly to a Poisson model, thus affecting its performance on high quantile levels. The other Bayesian model, BETANBB, is frequently effective: its results are particularly good on the extreme quantiles for Auto, and in the central ones for all data sets, confirming the suitability of the negative binomial distribution on intermittent demand.

Bootstrapping methods sometimes rank in top positions, but they lack consistency across different metrics. VZ, for instance, performs well on high quantiles only when time series are long (Pasta, TinyM5). Its mean forecasts are accurate only on the Auto dataset, the one with the least sparse demand. Its main limitation is that, using Croston's decomposition, the number of values to sample from is equal to the number of non-zero values in the time series. Similarly, the performance of WSS is unstable, probably because bootstrap demand values, after the jittering step, are then rounded to positive integers: this latter step biases the quantile forecasts of the demand size upwards. Instead in RAF, where the lowest quantiles are only driven by the occurrence Markov Chain, WSS has good scores.

ARMA-based models are probably the class with the worst performance. NNARMA and MARWAL sometimes perform well on RMSSE, with the latter ranking second on RAF. Yet on probabilistic scores the two methods are at the bottom of
the ranking, particularly at the central quantiles where they rank 8th to 11th on every dataset.
This gap between a competitive mean forecast and poor-performing quantile forecasts can be traced back to the use of a Gaussian forecast distribution, whose symmetric tails fail to model frequent zeros and sporadic, possibly large, demand spikes.

\section{A worked example of intermittent time series reconciliation}
\label{sec:integration}


\begin{figure}[h!]
\centering
\includegraphics[width=\textwidth]{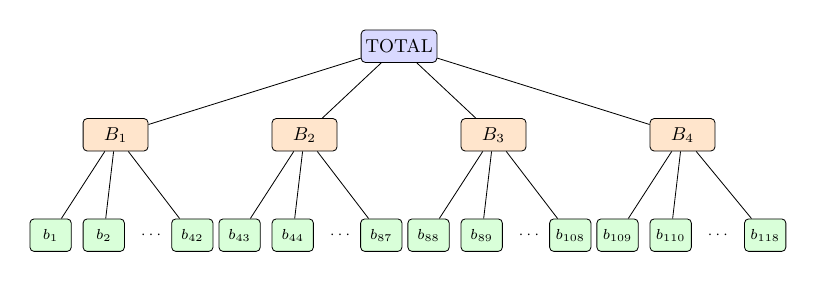}
\caption{Hierarchy of the Pasta dataset.}
\label{fig:hierarchy}
\end{figure}

The integration of \fableintermittent in the \pkg{fable} framework automatically gives access to time series reconciliation tools. Within the \pkg{fable} framework, in particular, \pkg{fable.bayesRecon} provides methods for the reconciliation of probabilistic forecasts of smooth, intermittent and mixed time series hierarchies.  
Here, we show how to produce coherent probabilistic forecasts \citep{Athanasopoulos_Hyndman_Kourentzes_Panagiotelis_2024} for
a hierarchy of time series. 
We use the pasta dataset, which has a natural hierarchical structure, with 118 bottom-level time series (daily sales of pasta products) aggregated into 4 time series representing the sales of each brand, further aggregated into total sales. The hierarchy is represented in Fig.~\ref{fig:hierarchy}. The code snippet in Listing~\ref{lst:data} shows the code for loading the data and creating the hierarchical structure and the train/test split. The experimental setup is the same as the one used in the previous section. 

\begin{listing}[h!]
\begin{minted}[
    linenos,
    numbersep=5pt,
    style=tango,
    bgcolor=quartobg,
    fontsize=\small,
    breaklines,
    frame=none,
    xleftmargin=1.5em,  % keeps numbers from crowding the background box edge
    escapeinside=@@
]{r}
library(fable)
library(@\fableintermittentname@)
library(dplyr)

# Total / brand / brand-product hierarchy (1 + 4 + 118 = 123 series)
  pasta_hts <- pasta |>
    aggregate_key(brand / product, value = sum(value))

  n_test <- 28

  train <- pasta_hts |> filter(index <= max(index) - n_test)
  test  <- pasta_hts |> filter(index >  max(index) - n_test)
\end{minted}
\caption{Loading the data and computing the aggregated series with \code{aggregate\_key}.}
\label{lst:data}
\end{listing}

The bottom time series are mostly non-smooth, as shown in Fig.~\ref{fig:sbc}; therefore we fit a TWEES model. On the other hand, the aggregated time series are smooth, so we fit an ETS model \citep[Ch.~8]{hyndman2021fpp3}. We reconcile these forecasts with the BUIS \citep{zambon2024efficient}, MixCond and TDcond \citep{Zambon_Azzimonti_Rubattu_Corani_2024} algorithms, implemented by the functions \code{bayesRecon\_BUIS()}, \code{bayesRecon\_mixCond()} and \code{bayesRecon\_TDcond()} of the package \pkg{fable.bayesRecon}.
As a comparison, we fit an ETS model to all the series of the hierarchy, which we then reconcile with MinT \citep{wickramasuriya2019optimal}.

\begin{listing}[h!]
\begin{minted}[
    linenos,
    numbersep=5pt,
    style=tango,
    bgcolor=quartobg,
    fontsize=\small,
    breaklines,
    frame=none,
    xleftmargin=1.5em  % keeps numbers from crowding the background box edge
]{r}
# Fit base forecasting models
# base_mix -> bottom TWEES, rest ETS
fit <- train_hts |>
  model(ets = ETS(value), twees = TWEES(value)) |>
  mutate(base_mix = if_else(is_aggregated(product), ets, twees))

# reconcile: BUIS, MixCond and TDcond on base_mix, MinT on ETS
fit_rec <- fit |> reconcile(BUIS    = bayesRecon_BUIS(base_mix),
                            MixCond = bayesRecon_MixCond(base_mix),
                            TDcond  = bayesRecon_TDcond(base_mix),
                            MinT    = min_trace(ets))

# Forecast 28 steps ahead
fc <- fit_rec |> forecast(h = n_test)
\end{minted}
\caption{Fit and reconcile mixed forecasts.
}
\label{lst:fitReconcile}
\end{listing}

\input{tables/recon_pasta_spl_MAE}

Listing~\ref{lst:fitReconcile} shows that we can easily fit all models and reconcile their forecasts with very few lines of code and an intuitive syntax.
Tab.~\ref{tab:reconpasta_spl_MAE} shows the $\mathrm{RMSSE}$ and $\mathrm{sQS}_\tau$ ($\tau \in \{0.5, 0.75, 0.835, 0.975, 0.995\}$) for the base forecasts (ETS and mixed) and the reconciled forecasts (BUIS, MixCond, TDcond and MinT). 
Among the base forecasts, mixed is consistently better than ETS, as TWEES is more effective than ETS on the intermittent bottom series.
Moreover, reconciliation via BUIS and MixCond improves the base forecasts on all the scores except for the two highest quantiles, while TDcond and MinT are less effective. Indeed, BUIS and MixCond are tailored for hierarchies with count-valued bottom forecasts; conversely, MinT relies on Gaussian base forecasts, which often are not effective on intermittent time series.

\section{Conclusion}
\label{sec:conclusions}

We present \fableintermittent, an \proglang{R} package that implements a broad set of probabilistic methods for intermittent time series within the \pkg{fable} framework.
Its purpose is to bring together, under one interface, methods that were previously scattered across different packages and programming languages, or had no publicly available implementation at all.
We also introduce TWEES, a new exponential smoothing model with a Tweedie predictive distribution.  Finally, we provide an efficient implementation of the Tweedie distribution in the separate \proglang{R} package \tweediedistr.
Since the package adopts the tidy syntax of the \textit{tidyverts} ecosystem, every method is fitted, forecast, evaluated and visualized through the same concise pipeline, and can be applied to large collections of series with a single call.
All the models are compatible with the rest of the ecosystem: they can be used together with other fable forecasting models, and their forecasts reconciled across a hierarchy using functions available in the same ecosystem.


Several directions remain open for future work.
The static distributions performed strongly and could be refined.
For example, the models might account for seasonality or give more weight to recent observations.
PARAMSD, which currently selects among candidate distributions with a log-likelihood criterion, could use a quantile score, since decisions typically rely on the upper tail of the predictive distribution.
Another direction is to extend some of the models to incorporate exogenous variables, such as promotions, which strongly affect demand.
%
Finally, we stress that our implementations are open-source and ready to be extended by users; we encourage contributions: the guide on the GitHub repository of the package\footnote{
\ifblind 
Blinded for anonymous submission
\else \url{https://github.com/StefanoDamato/fable.intermittent/commit/ee95d66947a49740dd0d0bac50ade369c6a572e3}
\fi} details how to build new models and integrate them in the package.

\bibliographystyle{elsarticle-harv}
\bibliography{refs}

\appendix

\section{Methods from the literature}\label{app:methods}

We describe here the methods from the literature implemented in \fableintermittent. We denote by $y_1, \dots, y_T$ the values of a time series with training set of length $T$. We use $h$ to denote the length of the forecast horizon. Forecasts are generated for the variables $Y_{T+1}, \dots, Y_{T+h}$, where we use capital letters to denote a random variable. 

Several of the methods rely on \emph{Croston's decomposition}
\citep{Croston_1972}, which we review here. Let $t_1 < t_2 < \dots < t_n$ be
the $n$ periods at which a positive value is observed, i.e.
$y_{t_i} > 0$. The \emph{demand sizes} are the positive values themselves,
$d_i = y_{t_i}$ for $i = 1, \dots, n$; the \emph{demand intervals}
$\ell_i = t_i - t_{i-1}$ (with $\ell_0 = 0$) are the
number of periods between consecutive positive demands. Alternatively, the \emph{occurrence} process is used, that is the
binary time series $o_t = \mathbbm{1}_{\{y_t > 0\}}$ for $t = 1, \dots, T$.

\subsection{Bootstrapping methods}

\paragraph{WSS \citep{Willemain_Smart_Schwarz_2004}} This bootstrapping method uses a Markov chain with states $\{0, 1\}$ to model the occurrence. Denoting $\{O_i\}$ the Bernoulli variables of the Markov chain, the transition probabilities are estimated using the training set as 
\[
\Prob(O_{i+1} = j | O_i = k) = \frac{\sum_{i=1}^{T-1} \mathbbm{1}_{\{o_i = k, o_{i+1} = j\}}}{\sum_{i=1}^{T} \mathbbm{1}_{\{o_i = k\}}}.
\]
The positive part of the forecast distribution is obtained sampling demand sizes $d_i$ and adding a Gaussian noise with variance $\sqrt{d_i}$; the demand size is then rounded and negative values are set to 0. 

Thus, the forecast distribution is a mixture of a mass in zero and a bootstrap distribution on positive integers. Forecasts are generated via autoregressive sampling.

\paragraph{VZ \citep{Zhou_Viswanathan_2011}} This method is based on Croston's decomposition. Predictions are generated with independent bootstrap samples of demand intervals and demand size: the former determine when positive values will appear, the latter determine their magnitude. Training times are almost immediate.

\subsection{Bayesian models}

\paragraph{BETANBB \citep{Harvey_Fernandes_1989}} 

A Bayesian model using a negative binomial forecast distribution, that is $Y_i \sim \rmNB(r, p_i)$. The count parameter $r$ is estimated globally across the time series, while the probability parameter $p_i$ follows a Beta distribution, $p_i \sim \rmBeta(a_i, b_i)$, whose parameters are updated following the equations
\begin{equation}\label{eq:bayesian}
    a_{i+1} = \omega (a_i + v) + (1 - \omega), \quad b_{i+1} = \omega (b_i + y_{i})
\end{equation}
with $\omega \in (0, 1]$ and $v > 0$. The parameters $r$, $v$, $\omega$, $a_1$, $b_1$ are estimated to minimize the negative log-likelihood of the observations. Probabilistic forecasts are generated via autoregressive sampling.

\paragraph{GAMPOISB \citep{Harvey_Fernandes_1989}}

In this model, a hierarchical Gamma-Poisson structure is used: $Y_i \sim \rmPois(\lambda_i)$ with $\lambda_i ~ \sim \Gamma(a_i, b_i)$. Similarly to Eq.~\eqref{eq:bayesian}, the parameters of the distribution vary as 
\begin{equation*}
    a_{i+1} = \omega a_i + y_i, \quad b_{i+1} = \omega b_i + 1 
\end{equation*}
with $\omega \in (0, 1]$. Initial parameters, as $a_1$ and $b_1$, and the updating coefficient $\omega$ are learned via an optimizer to minimize the negative log-likelihood, which takes the form of a negative binomial distribution. One step-ahead forecasts are computed in closed-form, while for longer horizons autoregressive sampling is needed.

\subsection{Exponential smoothing models}

\paragraph{HSPES \citep{Snyder_Ord_Beaumont_2012}}
In this model, demand size and occurrence are forecast separately. To both, a simple exponential smoothing \citep[Ch.~8]{hyndman2021fpp3} is applied, such that respectively
\begin{equation*}
    \pi_i = \alpha_\pi o_{i-1} + \theta_\pi \bar{\pi} + (1 - \alpha_\pi - \theta_\pi) \pi_{i-1}
\end{equation*}
and
\begin{equation*}
    \lambda_i = \alpha_\lambda (d_{i-1}-1) +  \theta_\lambda \bar{\lambda} + (1-\alpha_\lambda - \theta_\lambda) \lambda_{i-1}
\end{equation*}
where for both $\eta \in \{\pi, \lambda\}$, $\alpha_\eta \geq 0, \theta_\eta \geq 0$, and $\alpha_\eta + \theta_\eta < 1$. The exponential smoothing is called \textit{undamped} or \textit{damped} depending on whether $\theta_\eta$ is equal to 0 or not.
Note that the shifted demand size time series $d_i - 1$ is potentially shorter than $y_1, \dots, y_T$ and is only updated when $y_i > 0$. 

The two processes are combined into a hurdle-shifted Poisson distribution (HSP) such that $Y_i \sim O_i (1 + Z_i)$ where $O_i \sim \rmBer(\pi_i)$ and $Z_i \sim \rmPois(\lambda_i)$. Thus $\Prob(Y_i=0)=1-\pi_i$ and $\Prob(Y_i = k) = \pi_i e^{-\lambda_i} \frac{\lambda_i^{k-1}}{(k-1)!}$.

The parameters of the exponential smoothing process, $\alpha_\eta$ and $\theta_\eta$, and the initial values $\eta_0$ for $\eta \in \{\pi, \lambda\}$ are learned to minimise the negative log-likelihood, while $\bar{\pi}$ and $\bar{\lambda}$ are set as the sample mean of the occurrence and the shifted demand size respectively.

\paragraph{NEGBINES \citep{Snyder_Ord_Beaumont_2012}}

In this model, a damped or undamped simple exponential smoothing process determines the mean parameter of a negative binomial distribution, which is related to the parametrisation as $\mu = r \frac{1-p}{p}$. Thus, $Y_i \sim \rmNB(\mu_i \frac{p}{1-p}, p)$, where the mean is computed with the recursion
\[
\mu_i = \alpha y_{i-1} + \theta \bar{\mu} + (1 - \alpha - \theta) \mu_{i-1}.
\]
The parameters $p \in (0, 1)$ and $\mu_1$ of the negative binomial distribution, and $\alpha \geq 0$ and $\theta \geq 0$ such that $\alpha + \theta < 1$ are determined using an optimiser. $\bar{\mu}$ is the mean of the training set of the time series.

The forecast distribution is available in closed form only for $h=1$; for longer horizons, it is obtained via autoregressive sampling. Being the negative binomial overdispersed, this model can expand its right tail more than HSPES; however, it is necessarily unimodal.

\subsection{Static methods}

\paragraph{EMPSD \citep{Boylan_Babai_2022}}
 The empirical distribution is a widely used and simple probabilistic benchmark for intermittent time series \citep{Spiliotis_Makridakis_Kaltsounis_Assimakopoulos_2021}. The predictive distribution is $\Prob(Y_i = k) = \frac1T\sum_{i=1}^T \mathbbm{1}_{\{ y_i = k\}}$.

\paragraph{PARAMSD \citep{Kolassa_2016}} Fitting static distributions has been shown to be a strong benchmark: \citet{Boylan_Babai_2022} show that, for slow-moving items, a well-chosen parametric distribution estimates the lead-time demand CDF as accurately as bootstrap resampling, particularly when the demand history is short. This model treats the series as i.i.d. draws from a count distribution: the predictive distribution is identical at every horizon, $Y_{T+j}\sim F_{\hat\Theta}$ for all $j$, where $\hat{\Theta}$ is a set of parameters learned via maximum likelihood. 
Five candidate distributions are fitted to the history $y_1,\dots,y_T$: Poisson ($\mathrm{Pois}(\lambda)$), negative binomial ($\mathrm{NB}(r,p)$), their hurdle-shifted counterparts \citep{feng2021comparison}, and Tweedie. 

Hurdle shifted models use a Croston-type decomposition into occurrence $o_t=\mathbbm{1}_{\{y_t>0\}}$ and positive demand sizes, placing an explicit mass $\hat\pi_0=1-\tfrac1T\sum_t o_t$ at zero and fitting the count distribution to the shifted sizes $\{y_t-1\mid y_t>0 \}$:
$$\mathbb P(Y=0)=\pi_0,\qquad \mathbb P(Y=k)=(1-\pi_0)p_{\Theta \setminus \{ \pi \}}(k-1),\quad k\ge 1 .$$
The choice of the distribution is driven by an information criterion (either the default $\mathrm{BIC} = -2\ell(\hat\Theta)+|\Theta|\log T$, or $\mathrm{AIC}=-2\ell(\hat\Theta)+2|\Theta|$, where $|\Theta|$ is the number of parameters),
and the minimiser is returned. 

To allow for a fair comparison with other distributions, we do not evaluate the log-density of the Tweedie distribution on count data, but discretise it by integrating it over the interval $[y - 1/2, y + 1/2)$ for any $y \in \mathbb{N}$. 

\subsection{ARMA methods}

\paragraph{MARWAL \citep{Sbrana_2025}} The model works as the interplay of an ARMA(1,1) process \citep{box1970time}, $z_i$, and a Markov chain $\chi_i$ on the occurrence switching between 0 and 1, such that
\[y_i = \chi_i \omega + z_i\] where $z_i = \lambda z_{i-1} + \nu_i + \theta \nu_{i-1}$ and $\omega$ is the mean of non-zero values of the training set.
The model achieves fast training times, as the parameters $\lambda$ and $\theta$ are estimated in closed form.

Furthermore, a validation
loop is performed to determine the starting value $t_0 \in \{1, \dots, T\}$ of the training set $y_{t_0}, \dots, y_T$.

The mean forecast $\hat{y}_{T+h}$ and its variance $\hat{\sigma}^2_{T+h}$ is computed in closed form for any horizon, and a Gaussian distribution is assumed for the forecasts such that
\[
Y_{T+h} \sim \mathcal{N} \left( \hat{y}_{T+h}, \hat{\sigma}^2_{T+h} \right).
\]
Negative samples or quantiles, however, are set to zero by design.

\paragraph{NNARMA \citep{Sbrana_Babai_2026}} 
This model drives the demand series through
a Gaussian ARMA(1,1) process, whose forecast distribution is then constrained
to remain non-negative. Writing
$\nu_i$ for the one step-ahead innovation, the model is
\begin{equation*}
 y_i - \omega = \phi (y_{i-1} - \omega) + \nu_i + \theta \nu_{i-1},
\end{equation*}
with $\phi, \omega > 0$, $\theta \le 0$ and $\phi + \theta > 0$. The three
parameters are estimated by least squares on the innovations $\nu_i$. The mean and variance of the forecast distribution
are computed in closed form for any horizon via the standard ARMA(1,1) recursion. 

Input time series undergo a multiplicative deseasonalisation, and seasonal factors, computed as the average ratio between the time series and its moving average, are then applied to the forecasts.
Similarly to the Markov Walk model, the distribution has its negative part set to zero. However, we underline that this leads to an ill-posed distribution, as collapsing the negative mass to zero would also modify its mean.

\section{Implementative differences between \tweediedistr and \pkg{tweedie}}\label{appendix:tweedie_algorithms}

\paragraph{\textup{\code{dtweedie()}}}
In \tweediedistr, thanks to the fast \proglang{C++} implementation, the density function \code{dtweedie()} can be evaluated via its series expansion \citep{Dunn_Smyth_2005} for any choice of parameters $\phi$ and $\rho \in (1,2)$ (the mean $\mu$ does not enter the series expansion). On the other hand, \pkg{tweedie} uses the series expansion only in a few cases. In the other cases, an interpolation strategy with saddlepoint approximation is used on tabulated data embedded in the package.

\paragraph{\textup{\code{ptweedie()}}} For the evaluation of the cumulative distribution function on positive values, the \pkg{tweedie} package uses the Fourier inversion strategy described by \citet{Dunn2008}.
For the \tweediedistr implementation, on the other hand, we use the Compound-Poisson parametrisation to write the cumulative distribution function as 
\begin{align*}
  \mathbb{P} (Y \leq y) = e^{-\lambda} + \sum_{n = 1}^{+ \infty} T_n(y) \ \text{with} \\ T_n(y) = p_{\mathrm{Pois}} (n \mid \lambda)F_\mathrm{Gamma}(y \mid n\alpha, \beta),
\end{align*}
for $y > 0$, where $p_{\mathrm{Pois}}$ denotes the mass function of a Poisson distribution and $F_\mathrm{Gamma}$ the cumulative distribution function of a Gamma distribution.
We then identify the index of the largest term
\begin{equation*}
\label{eq:M_ptweedie}
  M = \arg \max_{n \geq 1} T_n (y)
\end{equation*}
and we approximate the series $\sum_{n = 1}^{+ \infty} T_n(y)$ as 

\begin{equation*}
  \sum_{n \in \Lambda} T_n(y) \ \text{with} \
\Lambda := \left \{ n \in \mathbb{N} :\frac{T_n (y) }{T_M(y)} \geq e^{-37} \right \}.
\end{equation*}
The threshold of $e^{-37} \simeq 8 \cdot 10^{-17}$ has been chosen as it guarantees double precision in a 64-bit floating point arithmetic \citep{Dunn_Smyth_2005}.

\paragraph{\textup{\code{qtweedie()}}}
The quantile function also differs among the two packages. The major reason for the speedup is that both packages exploit their implementation of  cumulative distribution function, which is faster in \tweediedistr. However, the \pkg{tweedie} package relies only on Brent's method \citep{Brent1971}, while \tweediedistr uses a \proglang{C++} implementation of Newton-Raphson algorithm, where the bisection method kicks in where convergence is not reached in few steps ($\sim4 \%$ of the times).

\paragraph{\textup{\code{rtweedie()}}}
Sampling from a Tweedie distribution for $\rho \in (1, 2)$ easily leverages the expression in Eq.~\eqref{eq:GammaPois}: first, draw $N$ as a $\rmPois (\lambda)$, then draw $y$ from $\rmGamma (n \alpha, \beta)$. Both packages use this strategy.

\section{Benchmarking speed and accuracy of a Tweedie implementation}\label{appendix:tweedie_exp}

We benchmark \tweediedistr's \code{dtweedie()}, \code{ptweedie()} and
\code{qtweedie()} against their namesake functions in the reference
\pkg{tweedie} package, on identical inputs, over a grid of dispersion
$\phi \in \{0.2, 0.6, \dots, 7.0\}$ (18 values), power $\rho \in \{1.1, 1.2,
\dots, 1.9\}$ (9 values).
For each $(\phi, \rho)$ pair, the experiment is repeated 200 times to average
out sampling variability; this number is up to 54\% smaller for the combinations of parameters where the quantile function gives errors. We do not iterate over different values of $\mu$, always set to 1, as it does not enter the series expansion of the density \citet{Dunn_Smyth_2005}.

For the density and  cumulative distribution function, for each repetition we draw a probability of a zero observation $p_0 \sim
\mathrm{Unif}(0.2, 0.8)$, then generate the test input of size $n$, $x_1, \dots, x_n$ as independent draws from a zero-inflated Gamma distribution, that is,
$X_i \overset{\mathrm{iid}}{\sim} O \cdot Z$,
with $O \sim \rmBer(p_0)$ and $Z \sim \rmGamma(1, 1)$; this is supposed to mimic the properties of Tweedie-generated data.
For the quantile function, we generate the input $q_1, \dots, q_n$ from $Q_i \overset{\mathrm{iid}}{\sim} \mathrm{Unif}(0,1)$.
Both packages are evaluated on this same input. Wall-clock time is measured with \pkg{microbenchmark} \citep{microbenchmark}, running 10 replications per
comparison; the reported speedup is the ratio of the reference package's
mean time to \tweediedistr's mean time.

To account for the dependence of the computational speedup on the sample size, we run these experiments for $n = 2, 10, 50, 200, 500, 2000$; the results shown in Fig.~\ref{fig:tweedie_speed} are generated using $n = 200$ for \code{dtweedie()} and \code{ptweedie()}, and $n = 10$ for \code{qtweedie()}.
We observe that the gap in the speedup increases with the sample size for the cumulative function and its inverse, with the new implementations being up to 1000 times faster than the original one. The difference in speed between the two versions, on the contrary, decreases with $n$ for the density; however, even for the largest sample size we use, 2000, our novel implementation remains from 1.5 to 20 times faster than the \pkg{tweedie} package implementation.

The correctness of \fableintermittent is verified by an automated test suite that checks its output against \pkg{tweedie} with a tolerance of $10^{-8}$. In the experiments grid described above, the differences in the density evaluation between the two packages were below the machine epsilon used in the experiments. For the cumulative probability function and the quantile function, the median absolute difference is again smaller than the machine epsilon; but the maximum difference is respectively in the order of $5\times10^{-2}$ and $7\times10^{-4}$. In the few cases where there is a disagreement among the two packages, Monte Carlo simulations confirmed \fableintermittent to be more accurate.

\end{document}

%% file: tables/methods_table.tex
\begin{table}[!ht]
\makebox[\linewidth][c]{
\begin{tabular}{lllc}
\toprule
Name & Family & Reference & Croston-based \\
\midrule
\rowcolor{black!6}  WSS & Bootstrapping & \citet{Willemain_Smart_Schwarz_2004} & \cmark \\
\rowcolor{black!6}  VZ & Bootstrapping & \citet{Zhou_Viswanathan_2011} & \cmark \\
  BETANBB & Bayesian & \citet{Harvey_Fernandes_1989} & \xmark \\
  GAMPOISB & Bayesian & \citet{Harvey_Fernandes_1989} & \xmark \\
\rowcolor{black!6}  HSPES & Exponential smoothing & \citet{Snyder_Ord_Beaumont_2012} & \cmark \\
\rowcolor{black!6}  NEGBINES & Exponential smoothing & \citet{Snyder_Ord_Beaumont_2012} & \xmark \\
  EMPSD & Static distribution & \citet{Boylan_Babai_2022} & \xmark \\
  PARAMSD & Static distribution & \citet{Kolassa_2016} & \xmark \\
\rowcolor{black!6}  MARWAL & ARMA & \citet{Sbrana_2025} & \cmark \\
\rowcolor{black!6}  NNARMA & ARMA & \citet{Sbrana_Babai_2026} & \xmark \\
\bottomrule
\end{tabular}
}
\caption{Methods from the literature implemented in the package, grouped by methodological family.}
\label{tab:methods}
\end{table}

%% file: tables/datasets.tex
\begin{table}[h]
\makebox[\linewidth][c]{
\begin{tabular}{llrrccrr}
\toprule
Dataset & Frequency & $N$ & $L$ & Period & Zeros (\%) & ADI & CV$^2$ \\
\midrule
\rowcolor{black!6} Auto & Monthly & 3000 &    24 & 2010-01 -- 2011-12 & 22 & 1.26 & 0.35 \\
Pasta & Daily &   118 & 1,798 & 2014-01 -- 2018-12 & 24 & 1.25 & 0.61 \\
\rowcolor{black!6} RAF & Monthly & 5000 &    84 & 1996-01 -- 2002-12 & 90 & 10.50 & 0.46 \\
TinyM5 & Daily &   280 & 1,913 & 2011-01 -- 2016-04 & 51 & 2.06 & 0.37 \\
\bottomrule
\end{tabular}
}
\caption{Overview of the datasets included in the package. $N$ and $L$ denote respectively the number of time
series in the data set and their length.
The last three
columns report the overall proportion of zero observations, the median
across series of the average inter-demand interval (ADI), and the median
squared coefficient of variation of the non-zero demand sizes ($\mathrm{CV}^2$).}
\label{tab:datasets}
\end{table}

%% file: tables/auto_spl_MAE.tex
\begin{table}[h!]
\makebox[\linewidth][c]{%
\begin{tabular}{lr@{\hspace{0.55em}}lr@{\hspace{0.55em}}lr@{\hspace{0.55em}}lr@{\hspace{0.55em}}lr@{\hspace{0.55em}}lr@{\hspace{0.55em}}l}
\toprule
Model & \multicolumn{2}{c}{RMSSE} & \multicolumn{2}{c}{$\mathrm{sQS}_{0.5}$} & \multicolumn{2}{c}{$\mathrm{sQS}_{0.75}$} & \multicolumn{2}{c}{$\mathrm{sQS}_{0.835}$} & \multicolumn{2}{c}{$\mathrm{sQS}_{0.975}$} & \multicolumn{2}{c}{$\mathrm{sQS}_{0.995}$} \\
\midrule
\rowcolor{black!6}  WSS & 0.883 & {\scriptsize\textcolor{black!45}{9th}} & 0.448 & {\scriptsize\textcolor{black!45}{8th}} & 0.421 & {\scriptsize\textcolor{black!45}{9th}} & 0.356 & {\scriptsize\textcolor{black!45}{9th}} & \textbf{\underline{0.117}} & {\scriptsize\textcolor{black!45}{1st}} & 0.044 & {\scriptsize\textcolor{black!45}{2nd}} \\
\rowcolor{black!6}  VZ & \underline{0.839} & {\scriptsize\textcolor{black!45}{3rd}} & 0.435 & {\scriptsize\textcolor{black!45}{4th}} & 0.409 & {\scriptsize\textcolor{black!45}{5th}} & 0.349 & {\scriptsize\textcolor{black!45}{5th}} & 0.138 & {\scriptsize\textcolor{black!45}{7th}} & 0.097 & {\scriptsize\textcolor{black!45}{10th}} \\
  BETANBB & 0.848 & {\scriptsize\textcolor{black!45}{7th}} & 0.440 & {\scriptsize\textcolor{black!45}{5th}} & 0.415 & {\scriptsize\textcolor{black!45}{7th}} & 0.352 & {\scriptsize\textcolor{black!45}{7th}} & \textbf{\underline{0.117}} & {\scriptsize\textcolor{black!45}{1st}} & \textbf{\underline{0.041}} & {\scriptsize\textcolor{black!45}{1st}} \\
  GAMPOISB & 0.916 & {\scriptsize\textcolor{black!45}{10th}} & 0.478 & {\scriptsize\textcolor{black!45}{10th}} & 0.473 & {\scriptsize\textcolor{black!45}{10th}} & 0.410 & {\scriptsize\textcolor{black!45}{10th}} & 0.144 & {\scriptsize\textcolor{black!45}{8th}} & 0.055 & {\scriptsize\textcolor{black!45}{6th}} \\
\rowcolor{black!6}  HSPES & \underline{0.840} & {\scriptsize\textcolor{black!45}{5th}} & 0.441 & {\scriptsize\textcolor{black!45}{6th}} & \underline{0.404} & {\scriptsize\textcolor{black!45}{2nd}} & 0.347 & {\scriptsize\textcolor{black!45}{4th}} & 0.150 & {\scriptsize\textcolor{black!45}{11th}} & 0.084 & {\scriptsize\textcolor{black!45}{9th}} \\
\rowcolor{black!6}  NEGBINES & \underline{0.840} & {\scriptsize\textcolor{black!45}{5th}} & \underline{0.433} & {\scriptsize\textcolor{black!45}{2nd}} & \underline{0.405} & {\scriptsize\textcolor{black!45}{3rd}} & \textbf{\underline{0.343}} & {\scriptsize\textcolor{black!45}{1st}} & \textbf{\underline{0.117}} & {\scriptsize\textcolor{black!45}{1st}} & 0.045 & {\scriptsize\textcolor{black!45}{4th}} \\
\rowcolor{black!6}  TWEES & \underline{0.839} & {\scriptsize\textcolor{black!45}{3rd}} & 0.442 & {\scriptsize\textcolor{black!45}{7th}} & 0.408 & {\scriptsize\textcolor{black!45}{4th}} & 0.345 & {\scriptsize\textcolor{black!45}{3rd}} & \textbf{\underline{0.117}} & {\scriptsize\textcolor{black!45}{1st}} & 0.044 & {\scriptsize\textcolor{black!45}{2nd}} \\
  EMPSD & \textbf{\underline{0.838}} & {\scriptsize\textcolor{black!45}{1st}} & \underline{0.434} & {\scriptsize\textcolor{black!45}{3rd}} & 0.409 & {\scriptsize\textcolor{black!45}{5th}} & 0.350 & {\scriptsize\textcolor{black!45}{6th}} & 0.146 & {\scriptsize\textcolor{black!45}{9th}} & 0.100 & {\scriptsize\textcolor{black!45}{11th}} \\
  PARAMSD & \textbf{\underline{0.838}} & {\scriptsize\textcolor{black!45}{1st}} & \textbf{\underline{0.432}} & {\scriptsize\textcolor{black!45}{1st}} & \textbf{\underline{0.403}} & {\scriptsize\textcolor{black!45}{1st}} & \textbf{\underline{0.343}} & {\scriptsize\textcolor{black!45}{1st}} & 0.123 & {\scriptsize\textcolor{black!45}{5th}} & 0.052 & {\scriptsize\textcolor{black!45}{5th}} \\
\rowcolor{black!6}  MARWAL & 0.852 & {\scriptsize\textcolor{black!45}{8th}} & 0.452 & {\scriptsize\textcolor{black!45}{9th}} & 0.417 & {\scriptsize\textcolor{black!45}{8th}} & 0.353 & {\scriptsize\textcolor{black!45}{8th}} & 0.134 & {\scriptsize\textcolor{black!45}{6th}} & 0.067 & {\scriptsize\textcolor{black!45}{8th}} \\
\rowcolor{black!6}  NNARMA & 1.611 & {\scriptsize\textcolor{black!45}{11th}} & 0.791 & {\scriptsize\textcolor{black!45}{11th}} & 0.584 & {\scriptsize\textcolor{black!45}{11th}} & 0.460 & {\scriptsize\textcolor{black!45}{11th}} & 0.146 & {\scriptsize\textcolor{black!45}{9th}} & 0.066 & {\scriptsize\textcolor{black!45}{7th}} \\
\bottomrule
\end{tabular}
}
\caption{Auto data set: RMSSE and $\mathrm{sQS}_\tau$ ($\tau = 0.5, 0.75, 0.835, 0.975, 0.995$).}
\label{tab:auto_spl_MAE}
\end{table}

%% file: tables/pasta_spl_MAE.tex
\begin{table}[h!]
\makebox[\linewidth][c]{%
\begin{tabular}{lr@{\hspace{0.55em}}lr@{\hspace{0.55em}}lr@{\hspace{0.55em}}lr@{\hspace{0.55em}}lr@{\hspace{0.55em}}lr@{\hspace{0.55em}}l}
\toprule
Model & \multicolumn{2}{c}{RMSSE} & \multicolumn{2}{c}{$\mathrm{sQS}_{0.5}$} & \multicolumn{2}{c}{$\mathrm{sQS}_{0.75}$} & \multicolumn{2}{c}{$\mathrm{sQS}_{0.835}$} & \multicolumn{2}{c}{$\mathrm{sQS}_{0.975}$} & \multicolumn{2}{c}{$\mathrm{sQS}_{0.995}$} \\
\midrule
\rowcolor{black!6}  WSS & \underline{0.762} & {\scriptsize\textcolor{black!45}{9th}} & \underline{0.411} & {\scriptsize\textcolor{black!45}{6th}} & \underline{0.425} & {\scriptsize\textcolor{black!45}{7th}} & 0.381 & {\scriptsize\textcolor{black!45}{8th}} & 0.153 & {\scriptsize\textcolor{black!45}{8th}} & \underline{0.054} & {\scriptsize\textcolor{black!45}{6th}} \\
\rowcolor{black!6}  VZ & \underline{0.741} & {\scriptsize\textcolor{black!45}{7th}} & \underline{0.404} & {\scriptsize\textcolor{black!45}{2nd}} & \underline{0.416} & {\scriptsize\textcolor{black!45}{2nd}} & \underline{0.370} & {\scriptsize\textcolor{black!45}{5th}} & \underline{0.147} & {\scriptsize\textcolor{black!45}{7th}} & \underline{0.052} & {\scriptsize\textcolor{black!45}{3rd}} \\
  BETANBB & \underline{0.738} & {\scriptsize\textcolor{black!45}{4th}} & \underline{0.411} & {\scriptsize\textcolor{black!45}{6th}} & \underline{0.430} & {\scriptsize\textcolor{black!45}{8th}} & \underline{0.380} & {\scriptsize\textcolor{black!45}{7th}} & \underline{0.158} & {\scriptsize\textcolor{black!45}{9th}} & \underline{0.066} & {\scriptsize\textcolor{black!45}{9th}} \\
  GAMPOISB & \underline{0.840} & {\scriptsize\textcolor{black!45}{11th}} & \underline{0.537} & {\scriptsize\textcolor{black!45}{11th}} & 0.556 & {\scriptsize\textcolor{black!45}{11th}} & 0.493 & {\scriptsize\textcolor{black!45}{11th}} & 0.207 & {\scriptsize\textcolor{black!45}{11th}} & 0.085 & {\scriptsize\textcolor{black!45}{10th}} \\
\rowcolor{black!6}  HSPES & \underline{0.733} & {\scriptsize\textcolor{black!45}{3rd}} & 0.436 & {\scriptsize\textcolor{black!45}{8th}} & \underline{0.422} & {\scriptsize\textcolor{black!45}{6th}} & \textbf{\underline{0.365}} & {\scriptsize\textcolor{black!45}{1st}} & 0.161 & {\scriptsize\textcolor{black!45}{10th}} & 0.091 & {\scriptsize\textcolor{black!45}{11th}} \\
\rowcolor{black!6}  NEGBINES & \textbf{\underline{0.730}} & {\scriptsize\textcolor{black!45}{1st}} & \textbf{\underline{0.403}} & {\scriptsize\textcolor{black!45}{1st}} & \underline{0.416} & {\scriptsize\textcolor{black!45}{2nd}} & \underline{0.366} & {\scriptsize\textcolor{black!45}{2nd}} & \textbf{\underline{0.135}} & {\scriptsize\textcolor{black!45}{1st}} & \underline{0.051} & {\scriptsize\textcolor{black!45}{2nd}} \\
\rowcolor{black!6}  TWEES & \underline{0.731} & {\scriptsize\textcolor{black!45}{2nd}} & \underline{0.407} & {\scriptsize\textcolor{black!45}{5th}} & \underline{0.417} & {\scriptsize\textcolor{black!45}{4th}} & \underline{0.367} & {\scriptsize\textcolor{black!45}{3rd}} & \underline{0.136} & {\scriptsize\textcolor{black!45}{2nd}} & \underline{0.053} & {\scriptsize\textcolor{black!45}{5th}} \\
  EMPSD & \underline{0.740} & {\scriptsize\textcolor{black!45}{5th}} & \underline{0.404} & {\scriptsize\textcolor{black!45}{2nd}} & \textbf{\underline{0.415}} & {\scriptsize\textcolor{black!45}{1st}} & \underline{0.369} & {\scriptsize\textcolor{black!45}{4th}} & \underline{0.146} & {\scriptsize\textcolor{black!45}{6th}} & \underline{0.052} & {\scriptsize\textcolor{black!45}{3rd}} \\
  PARAMSD & \underline{0.740} & {\scriptsize\textcolor{black!45}{5th}} & \underline{0.405} & {\scriptsize\textcolor{black!45}{4th}} & \underline{0.421} & {\scriptsize\textcolor{black!45}{5th}} & \underline{0.376} & {\scriptsize\textcolor{black!45}{6th}} & \underline{0.141} & {\scriptsize\textcolor{black!45}{3rd}} & \textbf{\underline{0.050}} & {\scriptsize\textcolor{black!45}{1st}} \\
\rowcolor{black!6}  MARWAL & 0.766 & {\scriptsize\textcolor{black!45}{10th}} & 0.473 & {\scriptsize\textcolor{black!45}{10th}} & 0.488 & {\scriptsize\textcolor{black!45}{10th}} & 0.415 & {\scriptsize\textcolor{black!45}{10th}} & \underline{0.141} & {\scriptsize\textcolor{black!45}{3rd}} & \underline{0.061} & {\scriptsize\textcolor{black!45}{7th}} \\
\rowcolor{black!6}  NNARMA & \underline{0.753} & {\scriptsize\textcolor{black!45}{8th}} & 0.447 & {\scriptsize\textcolor{black!45}{9th}} & 0.466 & {\scriptsize\textcolor{black!45}{9th}} & 0.402 & {\scriptsize\textcolor{black!45}{9th}} & \underline{0.143} & {\scriptsize\textcolor{black!45}{5th}} & 0.063 & {\scriptsize\textcolor{black!45}{8th}} \\
\bottomrule
\end{tabular}
}
\caption{Pasta data set: RMSSE and $\mathrm{sQS}_\tau$ ($\tau = 0.5, 0.75, 0.835, 0.975, 0.995$).}
\label{tab:pasta_spl_MAE}
\end{table}

%% file: tables/raf_spl_MAE.tex
\begin{table}[h!]
\makebox[\linewidth][c]{%
\begin{tabular}{lr@{\hspace{0.55em}}lr@{\hspace{0.55em}}lr@{\hspace{0.55em}}lr@{\hspace{0.55em}}lr@{\hspace{0.55em}}lr@{\hspace{0.55em}}l}
\toprule
Model & \multicolumn{2}{c}{RMSSE} & \multicolumn{2}{c}{$\mathrm{sQS}_{0.5}$} & \multicolumn{2}{c}{$\mathrm{sQS}_{0.75}$} & \multicolumn{2}{c}{$\mathrm{sQS}_{0.835}$} & \multicolumn{2}{c}{$\mathrm{sQS}_{0.975}$} & \multicolumn{2}{c}{$\mathrm{sQS}_{0.995}$} \\
\midrule
\rowcolor{black!6}  WSS & 0.722 & {\scriptsize\textcolor{black!45}{10th}} & \textbf{\underline{0.316}} & {\scriptsize\textcolor{black!45}{1st}} & 0.475 & {\scriptsize\textcolor{black!45}{6th}} & 0.533 & {\scriptsize\textcolor{black!45}{4th}} & 0.415 & {\scriptsize\textcolor{black!45}{6th}} & \textbf{\underline{0.209}} & {\scriptsize\textcolor{black!45}{1st}} \\
\rowcolor{black!6}  VZ & 0.741 & {\scriptsize\textcolor{black!45}{11th}} & 0.317 & {\scriptsize\textcolor{black!45}{8th}} & 0.510 & {\scriptsize\textcolor{black!45}{8th}} & 0.592 & {\scriptsize\textcolor{black!45}{9th}} & 0.492 & {\scriptsize\textcolor{black!45}{11th}} & 0.364 & {\scriptsize\textcolor{black!45}{10th}} \\
  BETANBB & 0.709 & {\scriptsize\textcolor{black!45}{7th}} & \textbf{\underline{0.316}} & {\scriptsize\textcolor{black!45}{1st}} & \textbf{\underline{0.474}} & {\scriptsize\textcolor{black!45}{1st}} & \textbf{\underline{0.532}} & {\scriptsize\textcolor{black!45}{1st}} & 0.417 & {\scriptsize\textcolor{black!45}{7th}} & 0.225 & {\scriptsize\textcolor{black!45}{5th}} \\
  GAMPOISB & 0.720 & {\scriptsize\textcolor{black!45}{9th}} & 0.383 & {\scriptsize\textcolor{black!45}{9th}} & 0.543 & {\scriptsize\textcolor{black!45}{9th}} & 0.590 & {\scriptsize\textcolor{black!45}{8th}} & 0.472 & {\scriptsize\textcolor{black!45}{10th}} & 0.375 & {\scriptsize\textcolor{black!45}{11th}} \\
\rowcolor{black!6}  HSPES & 0.703 & {\scriptsize\textcolor{black!45}{2nd}} & \textbf{\underline{0.316}} & {\scriptsize\textcolor{black!45}{1st}} & 0.475 & {\scriptsize\textcolor{black!45}{6th}} & 0.539 & {\scriptsize\textcolor{black!45}{7th}} & \textbf{\underline{0.396}} & {\scriptsize\textcolor{black!45}{1st}} & 0.241 & {\scriptsize\textcolor{black!45}{7th}} \\
\rowcolor{black!6}  NEGBINES & 0.704 & {\scriptsize\textcolor{black!45}{6th}} & \textbf{\underline{0.316}} & {\scriptsize\textcolor{black!45}{1st}} & \textbf{\underline{0.474}} & {\scriptsize\textcolor{black!45}{1st}} & 0.533 & {\scriptsize\textcolor{black!45}{4th}} & 0.407 & {\scriptsize\textcolor{black!45}{5th}} & 0.215 & {\scriptsize\textcolor{black!45}{2nd}} \\
\rowcolor{black!6}  TWEES & \textbf{\underline{0.702}} & {\scriptsize\textcolor{black!45}{1st}} & \textbf{\underline{0.316}} & {\scriptsize\textcolor{black!45}{1st}} & \textbf{\underline{0.474}} & {\scriptsize\textcolor{black!45}{1st}} & \textbf{\underline{0.532}} & {\scriptsize\textcolor{black!45}{1st}} & 0.406 & {\scriptsize\textcolor{black!45}{4th}} & 0.221 & {\scriptsize\textcolor{black!45}{3rd}} \\
  EMPSD & \underline{0.703} & {\scriptsize\textcolor{black!45}{2nd}} & \textbf{\underline{0.316}} & {\scriptsize\textcolor{black!45}{1st}} & \textbf{\underline{0.474}} & {\scriptsize\textcolor{black!45}{1st}} & \textbf{\underline{0.532}} & {\scriptsize\textcolor{black!45}{1st}} & 0.400 & {\scriptsize\textcolor{black!45}{3rd}} & 0.234 & {\scriptsize\textcolor{black!45}{6th}} \\
  PARAMSD & \underline{0.703} & {\scriptsize\textcolor{black!45}{2nd}} & \textbf{\underline{0.316}} & {\scriptsize\textcolor{black!45}{1st}} & \textbf{\underline{0.474}} & {\scriptsize\textcolor{black!45}{1st}} & 0.535 & {\scriptsize\textcolor{black!45}{6th}} & 0.398 & {\scriptsize\textcolor{black!45}{2nd}} & 0.221 & {\scriptsize\textcolor{black!45}{3rd}} \\
\rowcolor{black!6}  MARWAL & 0.703 & {\scriptsize\textcolor{black!45}{2nd}} & 0.536 & {\scriptsize\textcolor{black!45}{10th}} & 0.798 & {\scriptsize\textcolor{black!45}{10th}} & 0.749 & {\scriptsize\textcolor{black!45}{10th}} & 0.429 & {\scriptsize\textcolor{black!45}{8th}} & 0.309 & {\scriptsize\textcolor{black!45}{8th}} \\
\rowcolor{black!6}  NNARMA & 0.715 & {\scriptsize\textcolor{black!45}{8th}} & 0.549 & {\scriptsize\textcolor{black!45}{11th}} & 0.805 & {\scriptsize\textcolor{black!45}{11th}} & 0.754 & {\scriptsize\textcolor{black!45}{11th}} & 0.429 & {\scriptsize\textcolor{black!45}{8th}} & 0.309 & {\scriptsize\textcolor{black!45}{8th}} \\
\bottomrule
\end{tabular}
}
\caption{RAF data set: RMSSE and $\mathrm{sQS}_\tau$ ($\tau = 0.5, 0.75, 0.835, 0.975, 0.995$).}
\label{tab:raf_spl_MAE}
\end{table}

%% file: tables/tinym5_spl_MAE.tex
\begin{table}[h!]
\makebox[\linewidth][c]{%
\begin{tabular}{lr@{\hspace{0.55em}}lr@{\hspace{0.55em}}lr@{\hspace{0.55em}}lr@{\hspace{0.55em}}lr@{\hspace{0.55em}}lr@{\hspace{0.55em}}l}
\toprule
Model & \multicolumn{2}{c}{RMSSE} & \multicolumn{2}{c}{$\mathrm{sQS}_{0.5}$} & \multicolumn{2}{c}{$\mathrm{sQS}_{0.75}$} & \multicolumn{2}{c}{$\mathrm{sQS}_{0.835}$} & \multicolumn{2}{c}{$\mathrm{sQS}_{0.975}$} & \multicolumn{2}{c}{$\mathrm{sQS}_{0.995}$} \\
\midrule
\rowcolor{black!6}  WSS & 0.907 & {\scriptsize\textcolor{black!45}{11th}} & 0.620 & {\scriptsize\textcolor{black!45}{7th}} & 0.740 & {\scriptsize\textcolor{black!45}{7th}} & \underline{0.693} & {\scriptsize\textcolor{black!45}{7th}} & \underline{0.271} & {\scriptsize\textcolor{black!45}{5th}} & \underline{0.082} & {\scriptsize\textcolor{black!45}{4th}} \\
\rowcolor{black!6}  VZ & 0.904 & {\scriptsize\textcolor{black!45}{10th}} & \underline{0.605} & {\scriptsize\textcolor{black!45}{6th}} & 0.701 & {\scriptsize\textcolor{black!45}{5th}} & \underline{0.646} & {\scriptsize\textcolor{black!45}{5th}} & \textbf{\underline{0.235}} & {\scriptsize\textcolor{black!45}{1st}} & \textbf{\underline{0.071}} & {\scriptsize\textcolor{black!45}{1st}} \\
  BETANBB & \underline{0.841} & {\scriptsize\textcolor{black!45}{4th}} & \textbf{\underline{0.568}} & {\scriptsize\textcolor{black!45}{1st}} & \textbf{\underline{0.640}} & {\scriptsize\textcolor{black!45}{1st}} & \underline{0.613} & {\scriptsize\textcolor{black!45}{3rd}} & \underline{0.289} & {\scriptsize\textcolor{black!45}{8th}} & 0.143 & {\scriptsize\textcolor{black!45}{8th}} \\
  GAMPOISB & 0.882 & {\scriptsize\textcolor{black!45}{7th}} & \underline{0.596} & {\scriptsize\textcolor{black!45}{5th}} & 0.701 & {\scriptsize\textcolor{black!45}{5th}} & \underline{0.683} & {\scriptsize\textcolor{black!45}{6th}} & 0.364 & {\scriptsize\textcolor{black!45}{11th}} & 0.184 & {\scriptsize\textcolor{black!45}{9th}} \\
\rowcolor{black!6}  HSPES & \underline{0.837} & {\scriptsize\textcolor{black!45}{3rd}} & \underline{0.575} & {\scriptsize\textcolor{black!45}{3rd}} & \underline{0.644} & {\scriptsize\textcolor{black!45}{2nd}} & \textbf{\underline{0.606}} & {\scriptsize\textcolor{black!45}{1st}} & \underline{0.254} & {\scriptsize\textcolor{black!45}{3rd}} & 0.116 & {\scriptsize\textcolor{black!45}{7th}} \\
\rowcolor{black!6}  NEGBINES & \underline{0.835} & {\scriptsize\textcolor{black!45}{2nd}} & \underline{0.575} & {\scriptsize\textcolor{black!45}{3rd}} & \underline{0.651} & {\scriptsize\textcolor{black!45}{3rd}} & \underline{0.617} & {\scriptsize\textcolor{black!45}{4th}} & \underline{0.264} & {\scriptsize\textcolor{black!45}{4th}} & 0.101 & {\scriptsize\textcolor{black!45}{6th}} \\
\rowcolor{black!6}  TWEES & \textbf{\underline{0.830}} & {\scriptsize\textcolor{black!45}{1st}} & \underline{0.570} & {\scriptsize\textcolor{black!45}{2nd}} & \underline{0.651} & {\scriptsize\textcolor{black!45}{3rd}} & \underline{0.609} & {\scriptsize\textcolor{black!45}{2nd}} & \underline{0.239} & {\scriptsize\textcolor{black!45}{2nd}} & \underline{0.084} & {\scriptsize\textcolor{black!45}{5th}} \\
  EMPSD & 0.899 & {\scriptsize\textcolor{black!45}{8th}} & 0.631 & {\scriptsize\textcolor{black!45}{8th}} & 0.769 & {\scriptsize\textcolor{black!45}{9th}} & \underline{0.716} & {\scriptsize\textcolor{black!45}{9th}} & \underline{0.276} & {\scriptsize\textcolor{black!45}{6th}} & \underline{0.078} & {\scriptsize\textcolor{black!45}{2nd}} \\
  PARAMSD & 0.899 & {\scriptsize\textcolor{black!45}{8th}} & 0.632 & {\scriptsize\textcolor{black!45}{9th}} & 0.768 & {\scriptsize\textcolor{black!45}{8th}} & \underline{0.717} & {\scriptsize\textcolor{black!45}{10th}} & \underline{0.284} & {\scriptsize\textcolor{black!45}{7th}} & \underline{0.079} & {\scriptsize\textcolor{black!45}{3rd}} \\
\rowcolor{black!6}  MARWAL & \underline{0.854} & {\scriptsize\textcolor{black!45}{6th}} & 0.668 & {\scriptsize\textcolor{black!45}{11th}} & 0.806 & {\scriptsize\textcolor{black!45}{11th}} & 0.730 & {\scriptsize\textcolor{black!45}{11th}} & 0.351 & {\scriptsize\textcolor{black!45}{10th}} & 0.207 & {\scriptsize\textcolor{black!45}{11th}} \\
\rowcolor{black!6}  NNARMA & \underline{0.844} & {\scriptsize\textcolor{black!45}{5th}} & 0.653 & {\scriptsize\textcolor{black!45}{10th}} & 0.776 & {\scriptsize\textcolor{black!45}{10th}} & 0.703 & {\scriptsize\textcolor{black!45}{8th}} & 0.338 & {\scriptsize\textcolor{black!45}{9th}} & 0.200 & {\scriptsize\textcolor{black!45}{10th}} \\
\bottomrule
\end{tabular}
}
\caption{TinyM5 data set: RMSSE and $\mathrm{sQS}_\tau$ ($\tau = 0.5, 0.75, 0.835, 0.975, 0.995$).}
\label{tab:tinym5_spl_MAE}
\end{table}

%% file: tables/recon_pasta_spl_MAE.tex
\begin{table}[h!]
\makebox[\linewidth][c]{%
\begin{tabular}{lr@{\hspace{0.55em}}lr@{\hspace{0.55em}}lr@{\hspace{0.55em}}lr@{\hspace{0.55em}}lr@{\hspace{0.55em}}lr@{\hspace{0.55em}}l}
\toprule
Model & \multicolumn{2}{c}{RMSSE} & \multicolumn{2}{c}{$\mathrm{sQS}_{0.5}$} & \multicolumn{2}{c}{$\mathrm{sQS}_{0.75}$} & \multicolumn{2}{c}{$\mathrm{sQS}_{0.835}$} & \multicolumn{2}{c}{$\mathrm{sQS}_{0.975}$} & \multicolumn{2}{c}{$\mathrm{sQS}_{0.995}$} \\
\midrule
\rowcolor{black!6}  Base (mixed) & \underline{0.741} & {\scriptsize\textcolor{black!45}{3rd}} & \underline{0.421} & {\scriptsize\textcolor{black!45}{3rd}} & \underline{0.433} & {\scriptsize\textcolor{black!45}{3rd}} & \underline{0.380} & {\scriptsize\textcolor{black!45}{3rd}} & \textbf{\underline{0.138}} & {\scriptsize\textcolor{black!45}{1st}} & \textbf{\underline{0.052}} & {\scriptsize\textcolor{black!45}{1st}} \\
\rowcolor{black!6}  Base (ETS) & 0.803 & {\scriptsize\textcolor{black!45}{6th}} & 0.510 & {\scriptsize\textcolor{black!45}{5th}} & 0.568 & {\scriptsize\textcolor{black!45}{5th}} & 0.492 & {\scriptsize\textcolor{black!45}{5th}} & 0.159 & {\scriptsize\textcolor{black!45}{5th}} & \underline{0.060} & {\scriptsize\textcolor{black!45}{5th}} \\
  BUIS & \textbf{\underline{0.733}} & {\scriptsize\textcolor{black!45}{1st}} & \textbf{\underline{0.413}} & {\scriptsize\textcolor{black!45}{1st}} & \textbf{\underline{0.419}} & {\scriptsize\textcolor{black!45}{1st}} & \textbf{\underline{0.367}} & {\scriptsize\textcolor{black!45}{1st}} & \underline{0.139} & {\scriptsize\textcolor{black!45}{2nd}} & \underline{0.057} & {\scriptsize\textcolor{black!45}{2nd}} \\
  MixCond & \underline{0.735} & {\scriptsize\textcolor{black!45}{2nd}} & \textbf{\underline{0.413}} & {\scriptsize\textcolor{black!45}{1st}} & \underline{0.420} & {\scriptsize\textcolor{black!45}{2nd}} & \underline{0.368} & {\scriptsize\textcolor{black!45}{2nd}} & \underline{0.142} & {\scriptsize\textcolor{black!45}{3rd}} & \underline{0.060} & {\scriptsize\textcolor{black!45}{5th}} \\
  TDcond & 0.796 & {\scriptsize\textcolor{black!45}{4th}} & \underline{0.432} & {\scriptsize\textcolor{black!45}{4th}} & \underline{0.442} & {\scriptsize\textcolor{black!45}{4th}} & 0.396 & {\scriptsize\textcolor{black!45}{4th}} & 0.159 & {\scriptsize\textcolor{black!45}{5th}} & \underline{0.058} & {\scriptsize\textcolor{black!45}{3rd}} \\
  MinT & 0.799 & {\scriptsize\textcolor{black!45}{5th}} & 0.510 & {\scriptsize\textcolor{black!45}{5th}} & 0.572 & {\scriptsize\textcolor{black!45}{6th}} & 0.494 & {\scriptsize\textcolor{black!45}{6th}} & 0.158 & {\scriptsize\textcolor{black!45}{4th}} & \underline{0.058} & {\scriptsize\textcolor{black!45}{3rd}} \\
\bottomrule
\end{tabular}
}
\caption{
RMSSE and $\mathrm{sQS}_\tau$ computed on the hierarchy of the Pasta dataset (Fig.~\ref{fig:hierarchy}).
Base forecasts (grey background) are incoherent; mixed are computed with TWEES on the bottom series and ETS on the aggregated series.   
}
\label{tab:reconpasta_spl_MAE}
\end{table}